\documentclass[11pt]{article}

\usepackage[preprint]{acl}

\usepackage{times}
\usepackage{latexsym}
\usepackage[ruled,vlined]{algorithm2e}
\usepackage{booktabs}
\usepackage{tabularx}
\usepackage{array}
\usepackage{multirow}
\usepackage{graphicx}
\usepackage[table]{xcolor}
\cellcolor{gray!20}
\usepackage{arydshln}
\usepackage{multirow}
\usepackage{amsmath}
\usepackage{arydshln} 

\usepackage{xcolor}
\usepackage[many]{tcolorbox}

\definecolor{BoxFrame}{HTML}{2C3E50}        
\definecolor{BoxBackground}{HTML}{F8FAFC}   
\definecolor{TitleBackground}{HTML}{2C3E50} 
\definecolor{TitleText}{HTML}{FFFFFF}       
\definecolor{PromptBlue}{HTML}{D6EAF8} 

\tcbset{
  academicbox/.style={
    boxsep=5pt,
    left=2pt,
    right=2pt,
    bottom=0.5pt,
    boxrule=0.5pt,
    colback=BoxBackground,      
    colframe=BoxFrame,
    colbacktitle=TitleBackground,
    coltitle=TitleText,
    enhanced,
    attach boxed title to top left={yshift=-0.1in,xshift=0.1in},
    boxed title style={boxrule=0pt,colframe=white},
    title={#1},
  }
}

\newtcolorbox{AcademicBox}[1][]{academicbox=#1}

\usepackage[T1]{fontenc}

\usepackage[utf8]{inputenc}

\usepackage{microtype}

\usepackage{inconsolata}

\usepackage{graphicx}
\usepackage{amsmath, amssymb, amsthm}
\usepackage{mathtools}

\title{Are Tools All We Need? Unveiling the Tool-Use Tax in LLM Agents}

\author{
 \textbf{Kaituo Zhang\textsuperscript{1}}\,\,
 \textbf{Zhen Xiong\textsuperscript{2,3}}\,\,
 \textbf{Mingyu Zhong\textsuperscript{1}}\,\,
 \textbf{Zhimeng Jiang\textsuperscript{4}}
\\
 \textbf{Zhouyuan Yuan\textsuperscript{5}}\,\,
 \textbf{Zhecheng Li\textsuperscript{5}}\,\,
 \textbf{Ying Lin\textsuperscript{1}}
\\
\\
 \textsuperscript{1}University of Houston\,\,
 \textsuperscript{2}University of Southern California\,\,
 \textsuperscript{3}New York University
\\
 \textsuperscript{4}Texas A\&M University\,\,
 \textsuperscript{5}University of California, San Diego
\\
}

\begin{document}
\maketitle
\begin{abstract}
Tool-augmented reasoning has become a popular direction for LLM-based agents, and it is widely assumed to improve reasoning and reliability. However, we demonstrate that this consensus does not always hold: in the presence of semantic distractors, tool-augmented reasoning does not necessarily outperform native CoT. To explain this performance gap, we propose a \textbf{Factorized Intervention Framework} that isolates the cost of prompt formatting, the overhead of the tool-calling protocol, and the actual gain from executing tools. 
Our analysis reveals a critical tradeoff: under semantic noise, the gains from tools often fail to offset the "tool-use tax", which is the performance degradation introduced by the tool-calling protocol itself. To address this, we introduce G-STEP, a lightweight inference-time gate to mitigate protocol-induced errors. While this yields partial recovery, our findings suggest that more substantial improvements still require strengthening the model’s intrinsic reasoning and tool-interaction capabilities.

\end{abstract}

\section{Introduction}

In recent years, augmenting large language models (LLMs) with external tools has become a central paradigm for extending reasoning and real-world task-solving capabilities. Function calling, search augmentation, and tool-augmented agents are widely regarded as effective ways to provide models with up-to-date information, precise computation, and access to external services \citep{zhang2026llmdataauditormetricoriented,chen-etal-2023-chatcot,liu2025toolacewinningpointsllm,chen-etal-2025-acebench,patil2025the}. Across recent progress in tool-use training and evaluation, a common assumption has emerged: tool use should improve performance and reliability.

However, this assumption has mostly been established under relatively clean input conditions. In realistic settings, model inputs often contain semantically related but reasoning-irrelevant context: information that is topically relevant, linguistically natural, and superficially plausible, yet unhelpful or misleading for the target reasoning chain. Recent studies show that such distractors can substantially disrupt reasoning-path selection and final accuracy, and that noisy external information may even amplify failures in reasoning and search-augmented systems \citep{xiong2025mappingmindsllmsgraphbased,yang2025llmreasoningdistractedirrelevant,lee2026lostnoisereasoningmodels,pham2026sealqaraisingbarreasoning}. These findings suggest that semantic distractors are a particularly revealing stress test for tool-conditioned reasoning, because they simultaneously challenge evidence selection, tool invocation, and the integration of external results.

At the same time, existing evaluations provide limited insight into this failure mode. Tool-use benchmarks mainly measure whether models can successfully call tools, while recent work on reasoning and agent evaluation increasingly argues that final-answer accuracy alone is insufficient for diagnosing intermediate failures \citep{chen-etal-2025-acebench,lee2025evaluatingstepbystepreasoningtraces,zhou2025dissectinglogicalreasoningllms,hwang2025assessingllmreasoningsteps,ou-etal-2025-agentdiagnose,winston2025taxonomy}. Yet these lines of work still stop short of explaining a basic question that arises in noisy semantic settings: \textit{why can native chain-of-thought (CoT) outperform a full tool-augmented protocol, even when tools are available in principle?} This gap is especially important for deployment, since failures in tool-augmented systems may stem not only from missing capabilities, but also from the mechanics of tool invocation and interaction itself \citep{cheng2026investigatingtoolmemoryconflictstoolaugmented,wang2026agentnoisebenchbenchmarkingrobustnesstoolusing}.

In this work, we study the CoT--Tool gap under semantic distractors through four sequential research questions: \textbf{whether} the gap emerges, \textbf{where} the expected gains from tool use are lost, \textbf{what} mechanism explains when tool use helps or hurts, and \textbf{whether} such failures can be mitigated by lightweight inference-time control. 
To answer these questions, we construct two benchmark datasets, GSM8K-Sem-Distractor and HotPotQA-Sem-Distractor, propose a factorized intervention framework that decomposes the gap into style cost ($\Delta_{sty}$), function-calling protocol overhead ($\Delta_{frc}$), and computation gain ($\Delta_{cmp}$), introduce the capability overlap principle to explain why tool gains are often redundant with capabilities already present in native CoT, and validate this account with a lightweight G-STEP intervention. 
Rather than arguing that tools are broadly ineffective, our goal is to identify when tool augmentation breaks down under semantically noisy conditions and to explain why, complementing recent work on inference-time correction and robustness \citep{tie2025llmscorrectthemselvesbenchmark}.

We summarize our contributions as follows:
\begin{itemize}




\item We introduce a CoT-centered view of tool-use tax, using native CoT as the reference point to analyze when tool augmentation provides marginal benefit or incurs additional cost under semantic distractors.

\item We propose a factorized intervention framework that decomposes the CoT–Tool gap into Function Calling (FC)-style formatting cost, tool-use protocol overhead, and real tool-execution gain.

\item We identify a capability-overlap pattern behind the CoT–Tool gap: many apparent tool gains occur on cases already solvable by native CoT, so redundant tool benefits may fail to offset protocol-induced failures.

\end{itemize}







\section{Methodology}

\subsection{Problem Setting: Semantic Distractor Stress Test}

\label{sec:dataset_description}

Most prior work evaluates models on clean benchmarks, whereas real-world inputs often contain semantically related noise. To better reflect this setting, we build a controllable augmentation pipeline that injects \emph{semantically relevant} context into benchmark samples; we call the resulting data \textbf{Sem-Distractor}.

We consider four distractor types (examples in Appendix~\ref{Sem-Distractor_Example}): Thematic Background (\textbf{TB}), which adds topic-related but logically unhelpful background; Semantic Paraphrase (\textbf{SP}), which paraphrases evidence while preserving meaning; Parallel Entity Distractor (\textbf{PED}), which introduces semantically similar but entity-confounding hard negatives; and Hedged Uncertainty (\textbf{HU}), which adds hedging cues such as ``reportedly'' to simulate uncertain information sources.

Rather than treating semantic distractors as generic data augmentation, we use them as a controlled stress test to probe whether tool-conditioned reasoning remains beneficial under semantically relevant but logically unhelpful noise. This setting allows us to study not only whether tool use helps, but also when the protocol itself becomes a source of degradation.


\subsection{A Factorized Intervention Framework}
\label{sec:method_framwork}

\begin{figure*}[t]
  \includegraphics[width=1\linewidth]{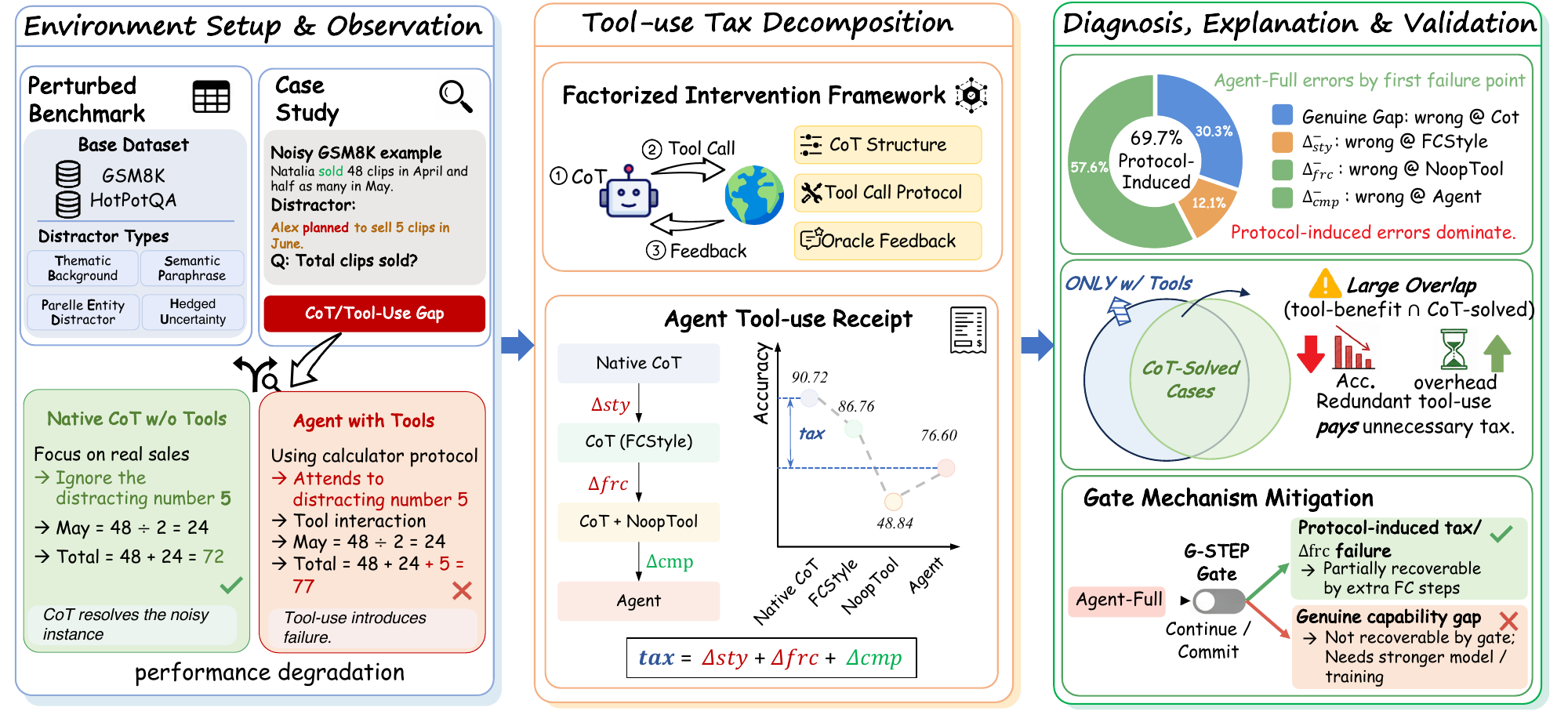}
\caption{Factorized intervention framework for diagnosing the CoT--Tool gap under semantic distractors. The framework identifies when tool-augmented reasoning underperforms native CoT, decomposes the gap into FC-style formatting cost, function-calling protocol overhead, and real tool-execution gain, and explains the gap through capability overlap and recoverability analysis.}
\label{fig:tool_variant}
\end{figure*}

We begin by comparing standard CoT and Agent-Full on the semantic distractor benchmarks, and define their end-to-end performance gap as the difference in final-answer accuracy. The direction and magnitude of this gap are determined empirically. However, this comparison only captures the net accuracy difference and doesn't reveal which components of the agentic pipeline drive the observed change. To disentangle these effects, we introduce the factorized intervention framework in Figure~\ref{fig:tool_variant}.

Figure~\ref{fig:tool_variant} summarizes our seven-condition intervention framework.
It consists of a primary decomposition chain from native reasoning to full tool-augmented reasoning:
\textbf{NoTool-CoT} $\rightarrow$ \textbf{NoTool-FCStyle} $\rightarrow$ \textbf{Agent-NoopTool} $\rightarrow$ \textbf{Agent-Full}.
Each transition adds one component of the tool-use pipeline: \textbf{NoTool-FCStyle} isolates the cost of adopting the function-calling prompt format without tool access; \textbf{Agent-NoopTool} adds the tool-use interaction protocol while replacing tool outputs with a no-op stub, estimating protocol overhead before useful tool execution; and \textbf{Agent-Full} restores real tool execution, measuring the gain from actual tool logic.

The remaining conditions serve as diagnostic probes around \textbf{Agent-Full}.
\textbf{Agent-OracleCalc} replaces tool outputs with the gold answer to bound computation-related losses;
\textbf{Agent-OracleEvidence} provides clean evidence to isolate distractor-sensitive evidence-selection failures;
and \textbf{Agent-Max1Turn} restricts the agent to a single FC turn to assess whether additional turns help or mainly add protocol overhead.
Detailed definitions of all seven conditions are provided in Appendix~\ref{app:condition_details}.

\subsection{Gap Decomposition Metrics}
\label{sec:decomp_framework}


\paragraph{Degradation chain and $\Delta$ decomposition.}
We define a degradation chain over four conditions that share the same noisy context but differ in protocol complexity:
\begin{equation}\label{eq:chain}
\text{CoT}
\;\xrightarrow{\;\Delta_{\text{sty}}\;}
\text{FCStyle}
\;\xrightarrow{\;\Delta_{\text{frc}}\;}
\text{NoopTool}
\;\xrightarrow{\;\Delta_{\text{cmp}}\;}
\text{Full}
\end{equation}
Here, $\mathrm{Acc}(x)$ denotes Accuracy. $\Delta_{\text{sty}} \triangleq \text{Acc(FCStyle)} - \text{Acc(CoT)}$ isolates the performance cost of strict FC-style prompt formatting in the absence of actual tool access. $\Delta_{\text{frc}} \triangleq \text{Acc(NoopTool)} - \text{Acc(FCStyle)}$ measures the function-calling protocol overhead incurred before useful tool execution can provide any gain. Finally, $\Delta_{\text{cmp}} \triangleq \text{Acc(Full)} - \text{Acc(NoopTool)}$ captures the net impact of executing real tool logic. Together, these components yield a strict additive decomposition of the total performance gap:
\begin{equation}\label{eq:identity}
\text{Acc(Full)} - \text{Acc(CoT)}
= \Delta_{\text{cmp}} + \Delta_{\text{frc}} + \Delta_{\text{sty}}
\end{equation}

We additionally define two auxiliary oracle probes for later bottleneck localization: $\Delta_{\text{oracle}} \triangleq \text{Acc(OracleCalc)} - \text{Acc(Full)}$, which upper-bounds computation-related losses, and $\Delta_{\text{context}} \triangleq \text{Acc(OracleEvid)} - \text{Acc(Full)}$, which isolates distractor-sensitive evidence-selection losses.


\subsection{Analytical Protocol for Root-Cause Diagnosis}
\label{sec:bridge}


We move from aggregate decomposition to finer-grained diagnosis of individual failures through two complementary views: a trajectory-level view, which characterizes how a failure manifests in the execution trajectory, and a sample-level view, which identifies where it first arises on the degradation chain. We therefore introduce a three-stage analytical protocol: (i) a trajectory-level failure taxonomy that identifies the dominant symptom of each failed trajectory, (ii) a sample-level attribution scheme that maps each failure to its earliest degradation point on the chain, and (iii) a capability-overlap analysis that measures whether tool-derived gains are redundant with the model's native reasoning ability.



\paragraph{Trajectory-Level Failure Taxonomy.}
\label{sec:failure_taxonomy}

For each incorrect Agent-Full trajectory, we assign a \emph{primary} failure label under a fixed priority order:
\begin{itemize}
\item \textbf{Type~A} (Under-computation): too few successful tool steps to complete the gold computation.
\item \textbf{Type~B} (Tool-execution error): at least one tool call fails or returns invalid output.
\item \textbf{Type~C} (Evidence drift): Evidence-F1 $< 0.5$, indicating substantial reliance on irrelevant context.
\item \textbf{Type~D} (Integration failure): the final tool output is correct, but the final prediction is not.
\item \textbf{Type~E} (No successful output): no tool call produces a usable result.
\item \textbf{Type~F} (Planning mismatch): the model follows a coherent but incorrect computation plan despite adequate evidence and functioning tools.
\end{itemize}

This taxonomy captures the \emph{observable symptom} of a failed trajectory rather than its root cause.

\paragraph{Sample-Level Attribution.}
\label{sec:sample_attr}

While Types~A--F characterize the surface symptom of a failed trajectory, they do not by themselves identify the principal source of degradation. We therefore further attribute each incorrect Agent-Full case to the earliest stage at which the sample becomes unsolved along the degradation chain in Eq.~\eqref{eq:chain}. This yields four mutually exclusive categories:
\begin{itemize}
\item \textbf{Genuine}: the sample is already incorrect under CoT;
\item \textbf{$\Delta_{\text{sty}}^{-}$}: the sample first becomes incorrect in FCStyle;
\item \textbf{$\Delta_{\text{frc}}^{-}$}: the sample first becomes incorrect in NoopTool;
\item \textbf{$\Delta_{\text{cmp}}^{-}$}: the sample remains correct up to NoopTool but becomes incorrect only in Agent-Full.
\end{itemize}

This attribution scheme complements the A--F taxonomy by separating failure symptom from degradation source. The former describes how an execution trace fails; the latter identifies where the failure is introduced in the sequence of controlled interventions.

\paragraph{Capability Overlap Analysis.}
\label{sec:capability_overlap}

To quantify the extent to which tool-derived gains provide capability beyond the model's native reasoning path, we identify \emph{tool-benefited} samples---cases where Agent-Full succeeds but the corresponding NoopTool control fails---and measure the fraction of such samples that are also solved by CoT. Let $\text{Full}(x)$, $\text{NoopTool}(x)$, and $\text{CoT}(x)$ denote binary correctness indicators for sample $x$ under the corresponding settings. We define
\[
\mathcal{B}_{\text{tool}} = \{x \mid \text{Full}(x)=1,\ \text{NoopTool}(x)=0\},
\]
and compute the overlap ratio as
\[
\text{Overlap} = \frac{|\{x \in \mathcal{B}_{\text{tool}} \mid \text{CoT}(x)=1\}|}{|\mathcal{B}_{\text{tool}}|}.
\]

A high overlap ratio indicates that many samples apparently helped by real tool execution are also solvable through the model's native reasoning path, suggesting substantial overlap between realized tool gains and the model's internal capability. This quantity therefore serves as a diagnostic signal for whether tool use contributes unique external capability or primarily recovers performance already achievable without external calls.

\subsection{Gate-Augmented Inference}
\label{sec:mitigation}


Building on the analysis from Section~\ref{sec:bridge}, we hypothesize that for tasks dominated by protocol-induced failures, especially premature termination or under-computation within the FC loop, a lightweight gate may help mitigate these errors.

\paragraph{Gate Design.}
We introduce \textsc{G-STEP}, a lightweight binary gate inserted at the termination point of the FC loop. When the model attempts to produce a final answer with no further tool calls, \textsc{G-STEP} decides whether to \emph{continue} tool-conditioned interaction or \emph{commit} to the current answer. The gate is trained to capture protocol-induced failures using \emph{CoT-fixability} as the primary supervision signal: cases where the FC agent fails but standard CoT succeeds are treated as instances where additional protocol-level intervention may still help. If the gate predicts continue, the system injects a continuation prompt and requires one additional tool-conditioned step before re-attempting the answer. We also evaluate a +\textsc{critic} variant, inspired by CRITIC~\citep{gou2024criticlargelanguagemodels}, which adds an explicit reflection step after calculator calls before the next tool action. This is motivated by our earlier finding that computation-chain errors dominate on GSM8K. Full details are provided in Appendix~\ref{Appendix:Gate_description}.


\section{Result \& Analysis}
\subsection{Experiment Settings}
This section summarizes the common experimental settings include the models, datasets, tool environments, and evaluation metrics. All prompts are provided in Appendix~\ref{Appendix:prompt}.

\paragraph{Models.}
We evaluate \textbf{Qwen3-4B}, \textbf{Qwen3-32B}, and \textbf{GPT-4.1-mini} as tool-enabled LLMs, covering open-source models at different scales and a closed-source counterpart. All models follow the same CoT, tool-augmented, and intervention protocols under a unified function-calling setup.

\paragraph{Datasets} Using the method in Section~\ref{sec:dataset_description}, we augment GSM8K~\citep{cobbe2021trainingverifierssolvemath} and HotpotQA~\citep{yang2018hotpotqadatasetdiverseexplainable} with GPT-4o-mini-generated distractors, yielding GSM8K-Sem-Distractor and HotpotQA-Sem-Distractor. 
For brevity, we refer to these two augmented benchmarks as GSM8K and HotpotQA in the rest of the paper, unless otherwise specified. 
Details are provided in Appendix~\ref{Appendix:Details_Semantic_Distractors}.

\paragraph{Tools \& Metrics} We use task-specific toolsets. For GSM8K-Sem-Distractor benchmark, the agent is equipped only with a calculator. For HotpotQA-Sem-Distractor benchmark, the agent is equipped with search sentences, read sentences, compare values, and calculator.

We evaluate performance with \textbf{Accuracy} and \textbf{Evidence-F1}.
\label{sec:results-rq}

\subsection{Semantic Distractor Stress Test}

\label{sec:Q1}
We evaluate Qwen3-4B, Qwen3-32B, and GPT-4.1-mini on GSM8K-Sem-Distractor under two settings: Agent-Full, which solves problems with tool use, and CoT, which relies on direct chain-of-thought reasoning. Implementation details of tool use are provided in Appendix~\ref{Apendix:fc_detail}.

\begin{table}[ht]
\centering
\renewcommand{\arraystretch}{1.25}
\resizebox{\columnwidth}{!}{%
\begin{tabular}{llcccccc}
\toprule
\textbf{Model} & \textbf{Setting} & \textbf{Base} & \textbf{TB} & \textbf{PED} & \textbf{HU} & \textbf{SP} & \textbf{Overall} \\
\midrule
\textsc{GPT-4.1-mini} & AF  & 75.40 & 76.00 & 76.60 & 77.80 & 77.20 & 76.60 \\
\textsc{GPT-4.1-mini} & CoT & \textbf{93.20} & \textbf{91.00} & \textbf{89.20} & \textbf{90.40} & \textbf{89.80} & \textbf{90.72} \\
\midrule
\textsc{Qwen3-4B}     & AF  & 51.20 & 54.20 & 48.00 & 57.40 & 49.60 & 52.08 \\
\textsc{Qwen3-4B}     & CoT & \textbf{91.00} & \textbf{86.80} & \textbf{81.40} & \textbf{83.60} & \textbf{84.40} & \textbf{85.44} \\
\midrule
\textsc{Qwen3-32B}    & AF  & 77.60 & 76.60 & 73.80 & 73.20 & 77.60 & 75.76 \\
\textsc{Qwen3-32B}    & CoT & \textbf{94.00} & \textbf{92.80} & \textbf{89.40} & \textbf{89.80} & \textbf{91.00} & \textbf{91.40} \\
\bottomrule
\end{tabular}%
}
\caption{Accuracy (\%) on GSM8K-Sem-Distractor. ``AF'' denotes Agent-Full. Bold values indicate the best performance.}
\label{tab:gsm8k_main_results}
\end{table}

Table~\ref{tab:gsm8k_main_results} shows a consistent pattern across all three models: CoT substantially outperforms Agent-Full on GSM8K-Sem-Distractor. The overall gaps are 14.12, 33.36, and 15.64\% for GPT-4.1-mini, Qwen3-4B, and Qwen3-32B, respectively. This result indicates that, in the presence of semantic distractors, adding tool use and multi-step agent protocols does not automatically improve robustness and can instead lead to substantial degradation.

Among distractor variants, PED is consistently one of the most challenging settings, particularly for CoT and the Qwen models. 
This indicates that entity-level semantic confusion is especially disruptive to evidence selection and downstream reasoning. 
Other distractor types show milder and less consistent effects, with only occasional small gains under Agent-Full for Qwen3-4B.

Overall, the CoT--Tool gap persists across models and distractor variants, indicating that tool-augmented reasoning can become vulnerable when semantic noise interacts with the tool-use protocol. We therefore turn to controlled interventions and mechanism-level analysis to explain where this degradation arises.

\subsection{Identify the CoT-Tool Gap from the Intervention Framework}
\label{sec:result_identify_gap}

\begin{table*}[bt]
\renewcommand{\arraystretch}{1.12}
\setlength{\tabcolsep}{4.2pt}
\centering
\small

\begin{tabular*}{\textwidth}{@{\extracolsep{\fill}} l ccc ccc @{}}
\toprule
\textbf{Condition} & \multicolumn{3}{c}{\textbf{GSM8K}} & \multicolumn{3}{c}{\textbf{HotPotQA}} \\
\cmidrule(lr){2-4} \cmidrule(lr){5-7}
& \textbf{Qwen3-4B} & \textbf{Qwen3-32B} & \textbf{GPT-4.1-mini} & \textbf{Qwen3-4B} & \textbf{Qwen3-32B} & \textbf{GPT-4.1-mini} \\
\midrule
NoTool-CoT       & 85.44 & 91.40 & 90.72 & 74.79 & 84.15 & 87.06 \\
NoTool-FCStyle   & 84.84 & 78.56 & 86.76 & 70.92 & 83.98 & 85.66 \\
Agent-NoopTool   & 30.64 & 50.92 & 48.84 & 56.69 & 82.07 & 84.87 \\
Agent-Full       & 52.08 & 75.76 & 76.60 & 72.32 & 83.03 & 86.44 \\
\cdashline{1-7}
Agent-Max1Turn   & 47.72 & 75.88 & 72.88 & 69.97 & 83.19 & 86.22 \\
Agent-OracleCalc & 89.20 & 93.40 & 82.24 & 93.05 & 90.70 & 91.71 \\
Agent-OracleEvid & 52.48 & 79.72 & 76.00 & 74.40 & 84.82 & 86.67 \\
\bottomrule
\end{tabular*}

\caption{Overall accuracy (\%) across seven experimental conditions. GSM8K uses exact-match accuracy. HotPotQA-32B and HotPotQA-GPT-4.1-mini use contains-match to accommodate free-form answers, while HotPotQA-4B uses exact match.}
\label{tab:all_conditions}
\end{table*}

Based on the Factorized Intervention Framework, Table~\ref{tab:all_conditions} reports overall Accuracy and Evidence-F1 across the seven experimental conditions. We analyze the average performance across all noisy variants. Detailed results for individual noise conditions are provided in Appendix~\ref{appendix:Results_in_Noisy_Condition}.


Three observations emerge from Table~\ref{tab:all_conditions}. 
First, \textbf{NoTool-CoT achieves the best non-oracle accuracy across all task--model pairs}, confirming that tool augmentation does not automatically improve robustness under semantic distractors. 
Second, the oracle probes suggest that the main bottleneck lies in the computation and tool-interaction chain rather than in evidence quality alone. 
For example, \textbf{Agent-OracleCalc} yields large gains over Agent-Full on Qwen3-4B, improving accuracy from 52.08\% to 89.20\% on GSM8K and from 72.32\% to 93.05\% on HotPotQA, whereas \textbf{Agent-OracleEvid} changes performance only marginally in comparison. 
Third, the degradation is strongly task-dependent: on GSM8K, Agent-Full trails CoT by 14.12--33.36\% across models, while on HotPotQA the gap is only 0.62--2.47\%. 
This contrast suggests that tool-use tax is amplified in sequential computation tasks, where errors introduced by the function-calling protocol can propagate through the reasoning chain, but is milder in retrieval-oriented QA settings where partial evidence or parametric knowledge may help compensate for imperfect tool interaction.

\paragraph{Decomposing the Performance Gap.}
Table~\ref{tab:delta_decomp} factorizes the aggregate CoT-to-Tool gap into its three components: $\Delta_{\text{cmp}}$, $\Delta_{\text{frc}}$, and $\Delta_{\text{sty}}$.




\begin{table}[t]
\centering
\small
\setlength{\tabcolsep}{4.5pt}

\begin{tabular}{@{}ll rrrr@{}}
\toprule
Task & Model
  & $\Delta_{\text{cmp}}$ & $\Delta_{\text{frc}}$
  & $\Delta_{\text{sty}}$ & Net \\
\midrule
\multirow{3}{*}{GSM8K}
  & 4B  & +21.44 & \textbf{$-$54.20} & $-$0.60  & $-$33.36 \\
  & 32B & +24.84 & \textbf{$-$27.64} & $-$12.84 & $-$15.64 \\
  & GPT & +27.76 & \textbf{$-$37.92} & $-$3.96  & $-$14.12 \\
\midrule
\multirow{3}{*}{HotPotQA}
  & 4B  & +15.63 & \textbf{$-$14.23} & $-$3.87  & $-$2.47 \\
  & 32B & +0.96  & \textbf{$-$1.91}  & $-$0.17  & $-$1.12 \\
  & GPT & +1.57  & $-$0.78            & \textbf{$-$1.40} & $-$0.62 \\
\bottomrule
\end{tabular}

\caption{$\Delta$ decomposition of the CoT-to-Tool gap (\%). Bold marks the largest-magnitude negative component in each row.}
\label{tab:delta_decomp}
\end{table}

This decomposition sharpens the counterintuitive finding: real tool execution can provide measurable gains, yet these gains may still be insufficient to offset the cost of entering the tool-use protocol. 
On GSM8K, $\Delta_{\mathrm{cmp}}$ is consistently positive, but $\Delta_{\mathrm{frc}}$ is the dominant negative component; for Qwen3-4B, for example, the protocol penalty is more than twice the tool-execution gain. 
This indicates that the problem is not the absence of useful computation, but that protocol overhead can corrupt more performance than tool execution recovers. 
HotPotQA presents a milder regime, where protocol costs are smaller and more easily offset by tool gains, leading to a much smaller CoT--Tool gap. 
Together, these results suggest that tool augmentation is not automatically beneficial: its value depends on whether realized tool gains are sufficiently complementary to overcome the surrounding protocol cost.

\paragraph{Bottleneck Localization.}
The oracle conditions show that computation quality---rather than evidence noise---is the primary limiting factor. Oracle computation yields large gains, whereas oracle evidence brings only modest improvements (Appendix~\ref{app:oracle_turn}). This motivates a closer examination of how errors arise within individual execution traces, which we address next.

\subsection{Decompose the CoT-Tool Gap}
\label{sec:decompose_cot_gap}

We next apply the diagnosis protocol from Section~\ref{sec:bridge} to distinguish genuine capability gaps from protocol-induced degradation. 
Each incorrect Agent-Full prediction is attributed to its earliest failure point along Eq.~\eqref{eq:chain}. 
Table~\ref{tab:delta_attr} summarizes the sample-level attribution, and Table~\ref{tab:af_cross} cross-tabulates these sources with the A--F failure taxonomy on GSM8K.


\begin{table}[t]
\centering
\small
\setlength{\tabcolsep}{4pt}

\begin{tabular}{@{}ll rrrrrr@{}}
\toprule
Task & Model & $N_{\text{w}}$
  & Gen. & $\Delta_{\text{sty}}^{-}$
  & $\Delta_{\text{frc}}^{-}$
  & $\Delta_{\text{cmp}}^{-}$
  & Proto. \\
\midrule
GSM8K
  & 4B  & 1198 & 20.6 & 11.4 & \textbf{58.7} & 9.3  & 79.4 \\
GSM8K
  & 32B & 606  & 24.1 & 24.9 & \textbf{45.5} & 5.4  & 75.8 \\
GSM8K
  & GPT & 585  & 30.3 & 12.1 & \textbf{44.6} & 13.0 & 69.7 \\
\midrule
HotPotQA
  & 4B  & 494  & \textbf{77.3} & 8.5  & 7.1 & 7.1 & 22.7 \\
HotPotQA
  & 32B & 303  & \textbf{73.3} & 7.6  & 13.2 & 5.9 & 26.7 \\
HotPotQA
  & GPT & 242  & \textbf{62.8} & 28.5 & 2.9 & 5.8 & 37.2 \\
\bottomrule
\end{tabular}

\caption{Sample-level $\Delta$ attribution (\%). ``Proto.'' = $\Delta_{\text{sty}}^{-} + \Delta_{\text{frc}}^{-} + \Delta_{\text{cmp}}^{-}$. $N_{\text{w}}$ denotes the number of incorrect samples, and Gen. denotes the proportion of genuinely incorrect samples. Bold marks the largest attributable source in each row.}
\label{tab:delta_attr}
\end{table}


\begin{table}[t]
\centering
\small
\setlength{\tabcolsep}{3.2pt}

\begin{tabular}{@{}ll rrrrr@{}}
\toprule
Model & Type & $N$ & Gen. & $\Delta_{\text{sty}}^{-}$
     & $\Delta_{\text{frc}}^{-}$ & $\Delta_{\text{cmp}}^{-}$ \\
\midrule
\multirow{4}{*}{4B}
& A Under-comp.  & 700 & 20.0 & 10.7 & \textbf{61.4} & 7.9 \\
& C Evid.~drift  & 281 & 23.5 & 13.2 & \textbf{54.1} & 9.3 \\
& D Integr.~fail & 11  & 0.0  & 9.1  & 36.4 & \textbf{54.5} \\
& F Plan.~mis.   & 195 & 19.5 & 11.8 & \textbf{56.4} & 12.3 \\
\midrule
\multirow{4}{*}{32B}
& A Under-comp.  & 423 & 22.9 & 23.9 & \textbf{47.8} & 5.4 \\
& C Evid.~drift  & 94  & 31.9 & 23.4 & \textbf{40.4} & 4.3 \\
& D Integr.~fail & 17  & 5.9  & 35.3 & \textbf{52.9} & 5.9 \\
& F Plan.~mis.   & 54  & 25.9 & 35.2 & \textbf{37.0} & 1.9 \\
\midrule
\multirow{4}{*}{GPT}
& A Under-comp.  & 297 & 35.7 & 11.1 & \textbf{42.4} & 10.8 \\
& C Evid.~drift  & 120 & 37.5 & 10.8 & \textbf{39.2} & 12.5 \\
& D Integr.~fail & 93  & 1.1  & 8.6  & \textbf{64.5} & 25.8 \\
& F Plan.~mis.   & 63  & 27.0 & 25.4 & \textbf{42.9} & 4.8 \\
\bottomrule
\end{tabular}

\caption{A--F $\times$ $\Delta$ cross-tabulation for GSM8K (row \%). Bold marks the dominant attributable source in each row. Rare types B and E are omitted here for compactness.}

\label{tab:af_cross}
\end{table}

Table~\ref{tab:delta_attr} reveals a robust task dichotomy. On GSM8K, a large majority of Agent-Full errors are protocol-induced: 79.4\% for Qwen3-4B, 75.8\% for Qwen3-32B, and 69.7\% for GPT-4.1-mini. In all three cases, $\Delta^{-}_{\mathrm{frc}}$ is the dominant source, accounting for 58.7\%, 45.5\%, and 44.6\% of all failures, respectively. By contrast, HotPotQA is dominated by genuine failures under which both CoT and the agent fail, comprising 77.3\% of errors for Qwen3-4B, 73.3\% for Qwen3-32B, and 62.8\% for GPT-4.1-mini. Protocol-induced errors therefore remain comparatively limited on HotPotQA, although GPT-4.1-mini exhibits a somewhat larger residual protocol component (37.2\%), driven primarily by $\Delta^{-}_{\mathrm{sty}}$ rather than $\Delta^{-}_{\mathrm{frc}}$.

Table~\ref{tab:af_cross} further shows that the A--F taxonomy captures failure symptoms rather than their underlying sources. 
On GSM8K, $\Delta^{-}_{\mathrm{frc}}$ is the dominant attribution for Types A, C, and F across all models, suggesting that apparent under-computation, evidence drift, and planning mismatch often arise after entering the function-calling protocol rather than from intrinsic reasoning deficits. 
This pattern is not limited to the Qwen models: for GPT-4.1-mini, $\Delta^{-}_{\mathrm{frc}}$ is also the largest contributor to Types A, C, and F, and even accounts for most Type~D errors. 
These results highlight why trajectory-level symptoms alone can be misleading: failures that appear to be planning or integration errors may instead be downstream effects of protocol-induced degradation.



This distinction is consistent with the oracle analysis (Appendix~\ref{Appendix:oracle_validation}): OracleCalc largely removes Types~A and~F, OracleEvid suppresses Type~C, and Type~D remains the main residual bottleneck once computation errors are corrected. Overall, A--F categories describe how a trajectory fails, whereas the $\Delta$ attribution identifies where that failure is introduced along the degradation chain.

The attribution analysis localizes where the losses arise, but it does not yet explain why the positive computation gain $\Delta_{\text{cmp}}$ still fails to outweigh the protocol overhead in aggregate.

\paragraph{The Capability Overlap Principle.}

This leads to a natural question: if real tool execution does provide useful computation, why does it still fail to offset the protocol tax? We answer this through capability overlap: the fraction of tool-benefited samples that are also solved by the model's native CoT path.


\begin{table}[t]
\centering
\small
\setlength{\tabcolsep}{3.6pt}
\renewcommand{\arraystretch}{1.08}

\begin{tabular*}{\columnwidth}{@{\extracolsep{\fill}}llrrr@{}}
\toprule
Task & Model & TB & CoT & Ovlp. (\%) \\
\midrule
\multirow{3}{*}{GSM8K}
  & 4B  & 673 & 603 & 89.6 \\
  & 32B & 669 & 629 & 94.0 \\
  & GPT & 780 & 744 & 95.4 \\
\multirow{3}{*}{HotPotQA}
  & 4B  & 332 & 292 & 88.0 \\
  & 32B & 58  & 39  & 67.2 \\
  & GPT & 57  & 32  & 56.1 \\
\bottomrule
\end{tabular*}

\caption{Capability overlap. TB denotes tool-benefited samples (Agent-Full correct and Agent-NoopTool wrong), CoT the subset of TB also solved by CoT, and Ovlp. the corresponding percentage. GPT denotes GPT-4.1-mini.}
\label{tab:overlap}
\end{table}


As shown in Table~\ref{tab:overlap}, capability overlap is extremely high on GSM8K across all three models: 89.6\% for Qwen3-4B, 94.0\% for Qwen3-32B, and 95.4\% for GPT-4.1-mini. In other words, most samples that appear to benefit from real tool execution are already solvable by the model's native CoT reasoning. Only 10.4\%, 6.0\%, and 4.6\% of the tool-benefited GSM8K cases are genuinely tool-essential.Thus, $\Delta_{\mathrm{cmp}}$ can be positive yet insufficient to offset the tool-use tax, because much of the realized tool gain overlaps with native CoT while $\Delta^{-}_{\mathrm{frc}}$ corrupts otherwise solvable examples.

HotPotQA exhibits a different pattern. Overlap remains high for Qwen3-4B (88.0\%), but drops to 67.2\% for Qwen3-32B and 56.1\% for GPT-4.1-mini. This indicates that capability overlap alone does not determine degradation severity: GPT-4.1-mini still shows only a small Full--CoT gap on HotPotQA despite substantially lower overlap, implying that the net effect also depends on the absolute protocol cost imposed by the task.


We summarize this pattern as the \textbf{Capability Overlap Principle}: 
\emph{when tool-provided capability substantially overlaps with the model's native reasoning ability, much of the realized tool gain becomes redundant with native CoT, while the tool-use protocol still incurs additional overhead. 
Under such conditions, a positive $\Delta_{\mathrm{cmp}}$ may still fail to offset the overall tool-use tax.}


\subsection{Cot-Tool Gap Mitigation}
\label{sec:result_gap_mitigation}

Table~\ref{tab:gate} reports the gate's performance across all four configurations. The gate is trained and evaluated on disjoint question splits, and all reported scores are measured on held-out test sets.\footnote{Baselines in Table~\ref{tab:gate} are re-evaluated on the gate test split and may differ slightly from the full-dataset values reported in Table~\ref{tab:all_conditions}.}


\begin{table}[hbt]
\centering
\small
\setlength{\tabcolsep}{4pt}
\renewcommand{\arraystretch}{1.08}

\begin{tabular}{@{}lrrrrrr@{}}
\toprule
Config & Full & Gate & +CRITIC & CoT & Gap & Cls. \\
\midrule
GSM-4B   & 50.64 & 69.12 & \textbf{74.88} & 82.64 & -32.00 & 75.75 \\
GSM-32B  & 73.28 & 77.04 & \textbf{77.44} & 91.20 & -17.92 & 23.21 \\
GSM-GPT  & 75.92 & 75.52 & \textbf{76.56} & 89.04 & -13.12 & 4.88 \\
\midrule
Hot-4B   & 73.18 & \textbf{74.97} & 74.53 & 76.87 & -3.69 & 48.51 \\
Hot-32B  & \textbf{83.02} & 82.90 & 82.10 & 84.15 & -1.13 & --- \\
Hot-GPT  & 86.37 & 85.81 & \textbf{87.04} & 87.15 & -0.78 & 85.72 \\
\bottomrule
\end{tabular}

\caption{Gate effectiveness on held-out test splits. Gap = Full -- CoT (pp). Cls.\ reports the fraction of the Full-to-CoT gap recovered by the better of Gate and +CRITIC. HotPotQA-32B and HotPotQA-GPT use contains-match; all others use exact match.}
\label{tab:gate}
\end{table}

\paragraph{Analysis.}

The gate's effectiveness closely follows the task dichotomy revealed by the $\Delta$ attribution analysis in Table~\ref{tab:delta_attr}: it yields substantial gains on GSM8K, where protocol-induced errors dominate, but only marginal or no benefit on HotPotQA, where most errors reflect genuine capability gaps. The strongest recovery appears on GSM8K-4B, where \textsc{G-STEP} improves accuracy from 50.64\% to 69.12\%, and the +\textsc{critic} variant further raises it to 74.88\%, closing 75.75\% of the Full-to-CoT gap. Gains are smaller on GSM8K-32B and HotPotQA-4B, and disappear on HotPotQA-32B, where errors are largely genuine rather than protocol-induced. This pattern suggests that gate-based intervention is most useful when failures are dominated by $\Delta_{\text{frc}}^{-}$-type protocol errors, whereas settings dominated by genuine capability gaps require stronger model improvements rather than additional inference-time control. GPT-4.1-mini follows the same qualitative trend: +CRITIC slightly improves over Agent-Full on both tasks, but the gain is modest on GSM8K and largely reflects closure of an already small residual gap on HotPotQA. Detailed analysis is provided in Appendix~\ref{app:gate_case_analysis}.



\paragraph{Robustness Across Distractor Types.}
The effectiveness of gate-based control is broadly consistent across the four semantic distractor categories. In particular, the gains are more pronounced in fragile GSM8K settings and remain limited on HotPotQA, which is consistent with the protocol-vs-capability dichotomy in Table~\ref{tab:delta_attr}. Detailed variant-wise robustness results are provided in Appendix~\ref{app:variant_robustness}.


\section{Related Work}

\subsection{Tool-augmented reasoning and function calling}


Recent research has explored tool augmentation and function calling as key directions for extending LLMs beyond language-only inference. ToolLLM~\citep{qin2024toolllm} introduced large-scale tool-use data construction, training, and evaluation over real-world APIs, while later studies improved tool learning through high-quality synthetic data~\citep{liu2025toolacewinningpointsllm}, structure-aware tool representation~\citep{su-etal-2025-toolscaler}, and token-level optimization with fine-grained error scoring~\citep{huang-etal-2025-ttpa}. In parallel, recent benchmarks have shifted toward realistic and process-oriented evaluation, covering fine-grained tool-use abilities~\citep{ye2024tooleyesfinegrainedevaluationtool} and mobile-assistant function calling under multi-turn, imperfect, and shifting user instructions~\citep{wang2025hammerbenchfinegrainedfunctioncallingevaluation}.

\subsection{Fine-Grained Diagnosis of Reasoning and Agent Failures}

Recent work increasingly argues that final-answer accuracy alone is insufficient for assessing reasoning quality. FineLogic~\citep{zhou2025dissectinglogicalreasoningllms}, Evaluating Step-by-step Reasoning Traces~\citep{lee2025evaluatingstepbystepreasoningtraces},~\citet{kim2025finalanswerevaluatingreasoning} all emphasize that intermediate reasoning traces reveal errors missed by outcome-only evaluation.

A similar shift has emerged in LLM agent evaluation. AgentDiagnose~\citep{ou-etal-2025-agentdiagnose},~\citet{zhang2025which}, AgentFail~\citep{ma2026demystifyinglifecyclefailuresplatformorchestrated}, and PaperArena~\citep{wang2026paperarenaevaluationbenchmarktoolaugmented} move beyond end-task success toward trajectory-level diagnosis, failure attribution, and root-cause analysis, a trend further summarized by recent survey work on LLM agent trajectory analysis~\citep{LLM_Agent_Trajectory_wang_2026}.

\section{Conclusion}

In this work, we studied the CoT--Tool gap under semantic distractors and showed that tool-augmented reasoning does not necessarily outperform native CoT. We analyzed this gap through a \textbf{Factorized Intervention Framework}, which decomposes tool gains and protocol costs into interpretable components, and further introduced the \textbf{Capability Overlap Principle}: many apparent tool gains arise on samples already solvable by native CoT, while the tool-calling protocol incurs broad additional overhead. We also showed that this analysis is actionable: a lightweight \textbf{G-STEP} intervention can partially mitigate protocol-induced failures, but its limited effect suggests that inference-time patching alone is insufficient for more substantial improvement. Overall, our findings suggest that tool augmentation is not universally beneficial: its value depends on whether tools provide genuinely complementary capability rather than redundant gains with added protocol burden.



\bibliography{custom}

\appendix

\section{Bottleneck Localization: Oracle Bounds and Multi-Turn Utility.}
\label{app:oracle_turn}
The oracle conditions explicitly identify where the remaining performance potential resides (Table~\ref{tab:oracle_turn}).

\begin{table}[htbp]
\centering\small
\setlength{\tabcolsep}{4pt}

\begin{tabular}{@{}ll rrr@{}}
\toprule
Task & Model & $\Delta_{\text{oracle}}$ & $\Delta_{\text{context}}$
     & $\Delta_{\text{turn}}$ \\
\midrule
\multirow{2}{*}{GSM8K}
  & 4B  & +37.1 & +0.4 & +4.4 \\
  & 32B & +17.6 & +4.0 & $-$0.1 \\
\multirow{2}{*}{HotPot}
  & 4B  & +20.7 & +2.1 & +2.4 \\
  & 32B & +7.7  & +1.8 & $-$0.2 \\
\bottomrule
\end{tabular}
\caption{Oracle upper bounds and multi-turn value (\%).}
\label{tab:oracle_turn}
\end{table}


For GSM8K--4B, the potential oracle gain 
($\Delta_{\text{oracle}} = +37.1\%$) is much larger than the context gain 
($\Delta_{\text{context}} = +0.4\%$), indicating that the main bottleneck lies in the computation chain rather than in evidence quality. 
For the 32B model, $\Delta_{\text{oracle}}$ remains substantial (+17.6\%), while $\Delta_{\text{context}}$ increases to +4.0\%, suggesting that evidence noise becomes more visible but still remains a secondary source of loss.

On HotPotQA, $\Delta_{\text{oracle}}$ is also non-negligible (+20.7\% for 4B and +7.7\% for 32B), showing that even in a retrieval-centric setting, tool-mediated answer production remains an important bottleneck. 
Across configurations, $\Delta_{\text{oracle}}$ is consistently larger than $\Delta_{\text{context}}$, suggesting that improving tool execution and computation quality is likely more beneficial than only denoising evidence.

Finally, multi-turn interaction exhibits a scale-dependent pattern: the 4B model benefits from additional interaction turns (+4.4\% on GSM8K and +2.4\% on HotPotQA), whereas the 32B model gains little 
($\Delta_{\text{turn}} \approx 0$ on both tasks). 
This suggests that additional turns are useful mainly for weaker models, while for stronger models they may introduce protocol overhead without substantial information gain.

\section{A--F Distribution and Oracle Validation.}
\label{Appendix:oracle_validation}
Table~\ref{tab:af_dist} reports the A--F distribution under Agent-Full and the two oracle conditions for GSM8K.

\begin{table}[ht]
\centering\small
\setlength{\tabcolsep}{3.5pt}
\begin{tabular}{@{}l rr rr rr@{}}
\toprule
 & \multicolumn{2}{c}{Agent-Full}
 & \multicolumn{2}{c}{OracleCalc}
 & \multicolumn{2}{c}{OracleEvid} \\
\cmidrule(lr){2-3}\cmidrule(lr){4-5}\cmidrule(lr){6-7}
Type & 4B & 32B & 4B & 32B & 4B & 32B \\
\midrule
A & 58.4 & 69.8 & 0 & 0 & 77.8 & 78.3 \\
C & 23.5 & 15.5 & 34.1 & 23.0 & 0 & 0 \\
D &  0.9 &  2.8 & 65.6 & 73.9 &  2.4 &  7.1 \\
F & 16.3 &  8.9 & 0 & 0 & 18.9 & 11.6 \\
\bottomrule
\end{tabular}
\caption{A--F failure distribution (\%) under Agent-Full and two oracle conditions on GSM8K.}
\label{tab:af_dist}
\end{table}

Under Agent-Full, Type~A is the dominant failure mode (58--70\%), followed by Type~C (16--24\%) and Type~F (9--16\%). A surface reading would attribute these errors to insufficient computation and motivate larger reasoning budgets. The oracle conditions show a different picture. OracleCalc eliminates Types~A and~F while leaving Type~C intact; OracleEvid eliminates Type~C while leaving Types~A and~F intact. Meanwhile, Type~D rises sharply under OracleCalc (66--74\%), indicating that once computation is made reliable, the remaining bottleneck is \emph{integration}: the agent fails to correctly consume and report tool outputs. This dissociation suggests that the A--F categories capture structurally distinct failure mechanisms.

\section{Detailed Condition Definitions}
\label{app:condition_details}

The seven experimental conditions differ in the specific component of the tool-use pipeline that is retained or intervened on.

In Agent-OracleCalc, the tool is still invoked, but directly returns the ground-truth answer, eliminating errors caused by malformed expressions or faulty tool use; the agent must still reason over the returned result and decide whether to output it. Agent-OracleEvidence provides an unperturbed, noise-free context to isolate the impact of distractors. Agent-Max1Turn does not impose a hard limit of one tool call; instead, it restricts the agent to a single protocol turn while still allowing multiple tool calls within that turn.

\begin{figure*}[ht]
  \includegraphics[width=1\linewidth]{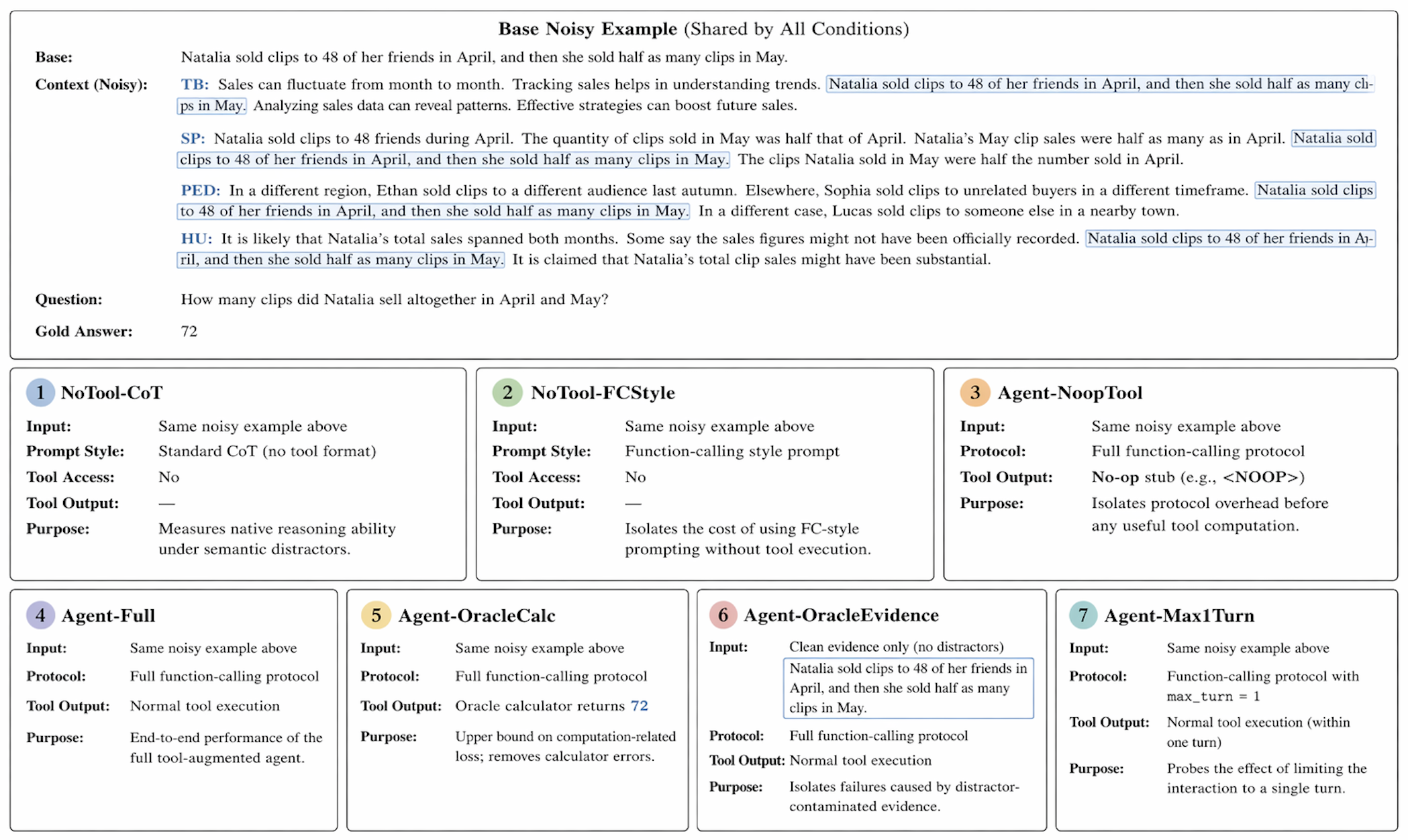}
\caption{
Illustrative example of the seven intervention conditions. 
All conditions share the same underlying Natalia clips question and gold answer, but differ in the component being intervened on: prompt style, tool-use protocol, tool output, evidence context, or interaction budget. 
The first four conditions form the primary decomposition chain from NoTool-CoT to Agent-Full, while Agent-OracleCalc, Agent-OracleEvidence, and Agent-Max1Turn serve as diagnostic probes for computation-related loss, distractor-sensitive evidence selection, and interaction-turn effects, respectively.
}
\label{fig:tool_using_example}
\end{figure*}

And we provide the example in Figure~\ref{fig:tool_using_example}.

\section{Secondary Patterns in the Decomposition Analysis}
\label{app:secondary_patterns}

Two secondary patterns further clarify these dynamics. First, the styling penalty is negligible for the 4B model on GSM8K ($-0.60$\%) but moderate on HotPotQA ($-3.87$\%), and becomes substantial for the 32B model on GSM8K ($-12.84$\%). This highlights a task-dependent interaction with model scale: on GSM8K, the 32B model's stronger reasoning chain is more vulnerable to FC formatting disruption ($-12.84$\% vs.\ $-0.60$\% for 4B), whereas on HotPotQA, where answers are shorter factoid spans, both models show minimal style sensitivity ($\leq 3.87$\%), with the 32B model being nearly unaffected ($-0.17$\%).

Second, the contrast in $\Delta_{\text{cmp}}$ across tasks is informative: on GSM8K, tools provide substantial but largely \emph{redundant} computational gains (+21 to +25\%) given the models' intrinsic mathematical ability; on HotPotQA, tools offer smaller but more \emph{genuine} information-acquisition utility, which helps explain the divergent net outcomes.

\section{Sem-Distractor Example}
\label{Sem-Distractor_Example}
In this section, we provide the examples of Sem-Distractor in Table~\ref{tab:noise_variants}.

\begin{table*}[t]
\centering
\small
\setlength{\tabcolsep}{5pt}
\renewcommand{\arraystretch}{1.15}
\begin{tabularx}{\textwidth}{>{\raggedright\arraybackslash}p{0.18\textwidth} >{\raggedright\arraybackslash}p{0.28\textwidth} >{\raggedright\arraybackslash}X}
\toprule
\textbf{Variant Type (Abbr.)} & \textbf{Perturbation Mechanism} & \textbf{Example from GSM8K} \\
\midrule
\textbf{Base} 
& The original, unmodified evidence sentence extracted directly from the dataset, serving as the ground truth. 
& ``Natalia sold clips to 48 of her friends in April, and then she sold half as many clips in May.'' \\

\textbf{Thematic Background (TB)} 
& Injects domain-relevant background knowledge using keywords from the question, providing contextually related but logically useless information. 
& ``Clips are commonly sold in bulk during seasonal events.'' \\

\textbf{Semantic Paraphrase (SP)} 
& Restates the original evidence to maintain semantic equivalence while altering the syntactic structure and vocabulary. 
& ``In April, Natalia's clip sales reached 48 units.'' \\

\textbf{Parallel Entity Distractor (PED)} 
& Introduces an alternative entity (person or scenario) performing a similar action to serve as a hard negative distractor. 
& ``Marcus sold clips to a different group of customers last summer.'' / ``Sarah sold clips during a separate event.'' \\

\textbf{Hedged Uncertainty (HU)} 
& Wraps the evidence in epistemic markers (e.g., \textit{reportedly}, \textit{some say}) to simulate an unreliable or unverified information source. 
& ``Natalia reportedly sold clips to a number of friends in April.'' / ``Some say the May sales might be about half.'' \\
\bottomrule
\end{tabularx}
\caption{Summary of semantic noise variants used in our data augmentation pipeline.}
\label{tab:noise_variants}
\end{table*}

\section{The Details of Semantic Distractors Dataset}
\label{Appendix:Details_Semantic_Distractors}

\subsection{The workflow of generating noisy dataset}
To demonstrate the workflow more clearly, we give the pseudo-code in Algorithm~\ref{alg:type_guided_noisy_augmentation}. 
\begin{algorithm}[t]
\caption{Type-Guided Noisy Data Augmentation}
\label{alg:type_guided_noisy_augmentation}
\KwIn{Base datasets $\mathcal{D}_{\text{gsm8k}}$ and $\mathcal{D}_{\text{hotpotqa}}$; distractor sentence number $k$; insertion positions $\mathcal{P}$; distractor type set $\mathcal{T}$; LLM generator $M=\texttt{gpt-4o-mini}$}
\KwOut{Augmented noisy dataset $\mathcal{D}_{\text{aug}}$}

$\mathcal{D}_{\text{aug}} \leftarrow \emptyset$\;

\ForEach{base sample $x \in \mathcal{D}_{\text{gsm8k}} \cup \mathcal{D}_{\text{hotpotqa}}$}{
    Determine the topic $t_x$ and keyword set $K_x$ of $x$\;
    Initialize distractor set $\Delta_x \leftarrow \emptyset$\;
    
    \For{$i \leftarrow 1$ \KwTo $k$}{
        Select distractor type $\tau_i \in \mathcal{T}$\;
        Generate distractor $d_i$ using $M$ conditioned on $(t_x, K_x, \tau_i)$\;
        $\Delta_x \leftarrow \Delta_x \cup \{d_i\}$\;
    }
    
    Insert $\Delta_x$ into $x$ according to positions $\mathcal{P}$ to obtain $\tilde{x}$\;
    $\mathcal{D}_{\text{aug}} \leftarrow \mathcal{D}_{\text{aug}} \cup \{\tilde{x}\}$\;
}

\Return{$\mathcal{D}_{\text{aug}}$}\;
\end{algorithm}

\subsection{The Examples of Noisy Dataset}

\label{Appendix:Example_Noisy_Dataset}

In this section, we present a set of examples from the generated noisy dataset, as illustrated in Figure~\ref{fig:Original-Generated}. In our work, the real dataset contains 22 distractor sentences. Therefore, the noise distribution is independent of the position.

\begin{figure*}[t]
    \begin{AcademicBox}[\footnotesize The Examples of Noisy Dataset]
    \textbf{Base: } \\
    Natalia sold clips to 48 of her friends in April, and then she sold half as many clips in May.\\
    
    \vspace{-5pt} \hrule \vspace{4pt}
    \textbf{Thematic Background (TB)} \\
    Sales can fluctuate from month to month. Tracking sales helps in understanding trends.\colorbox{PromptBlue} {Natalia sold clips to 48 of her friends in April, and then she sold half as many clips in May.} Analyzing sales data can reveal patterns. Effective strategies can boost future sales.\\
    
    \vspace{-5pt} \hrule \vspace{4pt}
    \textbf{Semantic Paraphrase (SP) } \\
    Natalia sold clips to 48 friends during April. The quantity of clips sold in May was half that of April. Natalia's May clip sales were half as many as in April. \colorbox{PromptBlue}{Natalia sold clips to 48 of her friends in April, and then she sold half as many clips in May.} The clips Natalia sold in May were half the number sold in April. In May, Natalia's clip sales were reduced to half of April's total.\\

        \vspace{-5pt} \hrule \vspace{4pt}
    \textbf{Parallel Entity Distractor (PED) } \\
    In a different region, Ethan sold clips to a different audience last autumn. Elsewhere, Sophia sold clips to unrelated buyers in a different timeframe. \colorbox{PromptBlue}{Natalia sold clips to 48 of her friends in April, and then she sold half as many clips in May.} In a different case, Lucas sold clips to someone else in a nearby town. Elsewhere, Isabella sold clips to a different group during a separate event.\\

        \vspace{-5pt} \hrule \vspace{4pt}
    \textbf{Hedged Uncertainty (HU) } \\
    It is likely that Natalia's total sales spanned both months. Some say the sales figures might not have been officially recorded. \colorbox{PromptBlue}{Natalia sold clips to 48 of her friends in April, and then she sold half as many clips in May.} It is claimed that Natalia's total clip sales might have been substantial. Some say the combined sales in April and May were possibly noteworthy.\\

    \vspace{-5pt} \hrule \vspace{4pt}
    \textbf{Question: } \\
    How many clips did Natalia sell altogether in April and May?\\

    \vspace{-5pt} \hrule \vspace{4pt}
    \textbf{Labels: } \\
    evidence sentence ids: [2]. noise sentence ids: [0, 1, 3, 4]. final: 72.\\

    \end{AcademicBox}
    \vspace{-1em}
    \caption{We provide one group of examples in noisy dataset. The groundtruth evidence has been highlighted using a \colorbox{PromptBlue}{blue box} for clarity. }
    \label{fig:Original-Generated}
\end{figure*}

\section{Tool-calling Agent Implementation Details}
\label{Apendix:fc_detail}
For all evaluated models, including GPT-4.1-mini and the Qwen series, we adopt the same function-calling(tool-calling) pipeline. GPT-4.1-mini is accessed through the OpenAI API, while Qwen models are served through vLLM. In both cases, the overall workflow remains the same:

\begin{enumerate}
    \item \textbf{Model layer:} The model decides whether to call a function and generates the corresponding call arguments.
    \item \textbf{Service layer:} The OpenAI API or vLLM \texttt{chat.completions} interface converts the model output into structured \texttt{tool\_calls}.
    \item \textbf{Execution layer:} Python code executes the actual function (e.g., a local calculator), and the returned result is fed back to the model for subsequent reasoning.
\end{enumerate}

This implementation therefore follows a unified and standard function-calling setup across different model families.

\section{Why high overlap leads to smaller degradation on HotPotQA}
\label{app:hotpot_overlap_discussion}

HotPotQA also exhibits substantial capability overlap. For Qwen3-4B, 88.0\% of tool-benefited cases are also solved by CoT, indicating that retrieval often returns information already recoverable from the model's parametric knowledge. However, despite this overlap, the net degradation on HotPotQA remains much smaller than on GSM8K (Table~\ref{tab:delta_decomp}). One likely reason is task structure. GSM8K requires multi-step sequential computation, where disrupting a single intermediate step can invalidate the entire reasoning chain. HotPotQA, by contrast, relies more on factoid retrieval and evidence aggregation, allowing the model to recover from imperfect tool interaction through partial evidence or parametric knowledge. Thus, overlap alone does not determine degradation severity; it interacts with the task's tolerance to protocol noise.

\section{G-STEP Design and Per-Configuration Analysis}
\label{Appendix:Gate_description}
\subsection{Stage 1: Data Collection}
The initial stage aims to obtain the raw performance data necessary to construct training labels. For each question (including multiple noise variants), the target Large Language Model (LLM) is evaluated under two independent conditions:

\begin{itemize}
    \item \textbf{Tool-Augmented Baseline ($\mathcal{D}_{\text{baseline}}$):} The model operates under the full function-calling (FC) protocol. For controlled intervention, the interaction is initialized with a required tool-use step, after which the model proceeds under a predefined maximum number of tool interactions. The model interacts with the environment to produce a final prediction.
    \item \textbf{Chain-of-Thought Baseline ($\mathcal{D}_{\text{CoT}}$):} The identical model receives the same question context but without access to any external tools. The model relies purely on internal CoT reasoning to produce a final prediction.
\end{itemize}

Both conditions operate on the exact same question set, enabling a per-sample comparison to identify instances where the model possesses the intrinsic reasoning capacity to solve a problem (via CoT) but fails when utilizing the tool-use protocol.

In this work, the gate is trained on a disjoint question set (GSM8K: 250 train / 250 test; HotPotQA: 178 train / 179 test), and performance metrics are reported strictly on the held-out test split.

\subsection{Stage 2: Label Construction and Feature Engineering}

\subsubsection{Label Construction}
For each sample, a binary label (\textit{continue} or \textit{commit}) is assigned. This label indicates whether the model should have continued utilizing tools after its initial tool call, based on hindsight knowledge derived from the gold answer and CoT performance. The priority rules for label assignment are detailed in Table \ref{tab:label_rules}.

\begin{table*}[htpb]
    \centering
    \begin{tabular}{cllp{6cm}}
        \toprule
        \textbf{Priority} & \textbf{Condition} & \textbf{Label} & \textbf{Rationale} \\
        \midrule
        1 & $\mathcal{D}_{\text{tool}}$ answered correctly & \textit{commit} & No intervention is required. \\
        2 & $\mathcal{D}_{\text{tool}}$ wrong, but $\mathcal{D}_{\text{CoT}}$ correct & \textit{continue} & Model reasoning is sufficient; the tool protocol caused the failure. \\
        3 & $\mathcal{D}_{\text{tool}}$ wrong, tool calls $< 2$ & \textit{continue} & Under-compute heuristic: premature stopping without sufficient exploration. \\
        4 & $\mathcal{D}_{\text{tool}}$ wrong, none of the above & \textit{commit} & Genuine reasoning failure; further continuation is unhelpful. \\
        \bottomrule
    \end{tabular}
    \caption{Priority rules for mixed-label construction. Priority 2 acts as the primary signal for the CoT-to-Tool gap.}
    \label{tab:label_rules}
\end{table*}

\subsubsection{Feature Engineering}
The intermediate state of the model after its first tool call is abstracted into a 120-dimensional, inference-safe numeric vector. The feature set comprises two main categories:
\begin{itemize}
    \item \textbf{Numeric Features (24 dimensions):} Captures execution progress (e.g., budget remaining), output stability, consistency between predictions and reasoning, reasoning signals (e.g., uncertainty indicators), evidence proxy, and tool trace properties.
    \item \textbf{Hashed Text Features (96 dimensions):} Captures textual patterns via the hashing trick (MD5 mapping to bin indices followed by $L_2$ normalization). The reasoning text yields a 64-dimensional vector, and the last tool output yields a 32-dimensional vector.
\end{itemize}

\subsection{Stage 3: Gate Training}
A binary classifier is trained to predict the probability of continuation, $P(\text{continue} \mid \text{state})$, given the 120-dimensional feature vector.

\begin{itemize}
    \item \textbf{Model Architecture:} Input features are standardized to zero-mean and unit-variance. The classifier is a two-layer Multilayer Perceptron (MLP) with 128 and 64 hidden units, utilizing ReLU activation functions.
    \item \textbf{Optimization Strategy:} The model is optimized using Adam with a learning rate of $10^{-3}$ and $L_2$ regularization ($\alpha = 10^{-4}$). Early stopping is applied using a 10\% validation split with a patience of 20 epochs.
    \item \textbf{Sample Weighting:} Labels are weighted by confidence levels derived from evidence specificity: strong ($3\times$), medium ($2\times$), and weak ($1\times$), with additional boosts applied for targeted error heuristics.
    \item \textbf{Cross-Validation:} A 5-fold GroupKFold validation is employed, grouped by question ID to strictly prevent information leakage across noise variants.
\end{itemize}

\subsection{Stage 4: Gate-Augmented Inference}
During the online inference phase, the trained gate is deployed dynamically inside the function-calling loop. When the model generates a final text without tool calls (attempting to submit), the gate evaluates the current state:

\begin{enumerate}
    \item Extract the 120-dimensional feature vector using the identical feature engineering protocol established during training.
    \item The MLP classifier outputs the probability $P(\text{continue})$.
    \item If $P(\text{continue}) \geq \tau$ (where the threshold $\tau$ is empirically set to $0.05$), the gate triggers a \textit{continue} decision. A continuation prompt (e.g., a CRITIC prompt forcing natural language reasoning prior to the next action) is injected into the context, forcing the model to initiate at least one additional tool call.
    \item If $P(\text{continue}) < \tau$, the gate decides to \textit{commit}, and the model's current answer is accepted as final.
\end{enumerate}

To ensure robustness, safety mechanisms including a maximum limit on extra turns (e.g., 3 extra turns), no-progress detection, and duplicate expression detection are implemented to prevent infinite generation loops.

\subsection{Training of Gate}
In this section, we provide the pseudocode for the training workflow of the gate in Algorithm~\ref{alg:gate_training}.

\begin{algorithm}[htpb]
    \caption{Offline Training Pipeline for Gate Decision MLP}
    \label{alg:gate_training}
    \KwIn{Baseline dataset $\mathcal{D}_{base}$, Chain-of-Thought dataset $\mathcal{D}_{cot}$}
    \KwOut{Trained Gate Decision Model $\mathcal{M}^*_{gate}$}
    
    \BlankLine
    \tcp{Phase 1: State Initialization \& Dataset Merging}
    $\mathcal{S} \leftarrow \text{PreprocessAndMerge}(\mathcal{D}_{base}, \mathcal{D}_{cot})$\;
    
    \BlankLine
    \tcp{Phase 2: Sample Annotation based on Heuristics}
    \ForEach{sample $s \in \mathcal{S}$}{
        \If{$s.\text{is\_baseline\_correct}$}{
            $y_s \leftarrow \text{commit}$\;
        }
        \ElseIf{$s.\text{is\_cot\_correct}$}{
            $y_s \leftarrow \text{continue}$\;
        }
        \ElseIf{$s.\text{tool\_calls} < 2$}{
            $y_s \leftarrow \text{continue}$\;
        }
        \Else{
            $y_s \leftarrow \text{commit}$\;
        }
    }
    $\mathcal{D}_{labeled} \leftarrow \{ (s, y_s) \}_{s \in \mathcal{S}}$\;
    
    \BlankLine
    \tcp{Phase 3: Feature Extraction \& Model Training}
    $\mathbf{X} \leftarrow \text{ExtractFeatures}(\mathcal{D}_{labeled}, d=120)$\;
    $\mathbf{y} \leftarrow \text{ExtractLabels}(\mathcal{D}_{labeled})$\;
    $\mathbf{X}_{norm} \leftarrow \text{StandardScaler}(\mathbf{X})$\;
    
    \BlankLine
    $\mathcal{M}_{gate} \leftarrow \text{InitializeMLP}(\text{hidden\_dims}=[128, 64])$\;
    $\mathcal{M}^*_{gate} \leftarrow \text{TrainWithGroupKFold}(\mathcal{M}_{gate}, \mathbf{X}_{norm}, \mathbf{y}, K=5)$\;
    
    \BlankLine
    \Return{$\mathcal{M}^*_{gate}$}
\end{algorithm}

\subsection{Inference with Gate}
In this section, we provide the pseudocode for the inference workflow with the gate in Algorithm~\ref{alg:online_inference}.
\begin{algorithm}[htpb]
    \caption{Online Inference Pipeline with Gate-Controlled Tool Calling}
    \label{alg:online_inference}
    
    \KwIn{Question $q$, Context $c$, Large Language Model $\mathcal{M}_{LLM}$, Trained Gate Model $\mathcal{M}_{gate}$}
    \KwIn{System prompt $P_{sys}$, Continue prompt $P_{cont}$, Probability threshold $\tau = 0.05$}
    \KwOut{Final answer $A$}
    
    \BlankLine
    \tcp{Initialize conversation history with system and user prompts}
    $\mathcal{H} \leftarrow P_{sys} \oplus \text{FormatUserPrompt}(q, c)$\;
    
    \BlankLine
    \While{\text{True}}{
        \tcp{Generate response from the LLM}
        $r \leftarrow \mathcal{M}_{LLM}(\mathcal{H})$\;
        
        \BlankLine
        \eIf{$\text{HasToolCall}(r)$}{
            \tcp{Execute tool and append observation to the loop}
            $o \leftarrow \text{ExecuteTool}(r.\text{tool\_call})$\;
            $\mathcal{H} \leftarrow \mathcal{H} \oplus r \oplus o$\;
        }{
            \tcp{Model attempts to submit final answer without tool calls}
            $p_{cont} \leftarrow \mathcal{M}_{gate}(\mathcal{H}, r)$ \tcp*{Compute $P(\text{continue})$}
            
            \BlankLine
            \eIf{$p_{cont} \ge \tau$}{
                \tcp{Continue the interaction by injecting the continue prompt}
                $\mathcal{H} \leftarrow \mathcal{H} \oplus r \oplus P_{cont}$\;
            }{
                \tcp{Accept the submission and halt}
                $A \leftarrow \text{ExtractFinalAnswer}(r)$\;
                \Return{$A$}\;
            }
        }
    }
\end{algorithm}

\subsection{Per-Configuration Analysis of Gate Effectiveness}
\label{app:gate_case_analysis}

The gate's behavior varies substantially across task--model settings, in ways that closely match the decomposition-based attribution analysis.

\paragraph{GSM8K-4B.}
GSM8K-4B is the most favorable setting for the gate, as it combines the largest performance gap ($-32.00$\%) with the highest $\Delta_{\text{frc}}^{-}$ share (58.7\% of all errors; Table~\ref{tab:delta_attr}). \textsc{G-STEP} alone improves accuracy from 50.64\% to 69.12\% (+18.48\%). Adding \textsc{critic} prompts further raises accuracy to 74.88\%, closing 75.75\% of the gap. This additional gain (+5.76\%) supports our earlier finding that GSM8K errors are primarily computation-chain failures (Types~A and~F): explicit reflection helps the weaker model identify arithmetic mistakes and produce better corrective tool-use strategies.

\paragraph{GSM8K-32B.}
GSM8K-32B provides a revealing contrast. Although protocol-induced errors remain prevalent overall (75.8\%; Table~\ref{tab:delta_attr}), gap closure reaches only 23.21\%. Two factors likely explain this. First, a much larger share of the 32B model's protocol-induced errors comes from $\Delta_{\text{sty}}^{-}$ (24.9\% vs.\ 11.4\% for 4B), i.e., format-sensitivity errors that additional tool turns cannot repair because the degradation occurs at the prompting stage rather than during iterative tool use. Second, even within the targeted $\Delta_{\text{frc}}^{-}$ errors, the 32B model appears less able to generate genuinely new corrective strategies in later turns. The minimal \textsc{critic} gain (+0.40\%) is consistent with this interpretation and with the observation in \S\ref{sec:bridge} that larger models are more sensitive to rigid format constraints than to invocation mechanics.

\paragraph{HotPotQA-4B.}
HotPotQA-4B presents a much narrower gap ($-3.69$\%), and 77.3\% of Agent-Full errors are genuine capability gaps (Table~\ref{tab:delta_attr}). The gate improves accuracy from 73.18\% to 74.97\%, a modest +1.79\% gain. Although this corresponds to 48.51\% gap closure, the absolute improvement is limited by the small protocol-induced error budget (22.7\%). The \textsc{critic} variant (74.53\%) performs slightly worse than the base gate, suggesting that on retrieval-heavy tasks, direct tool re-engagement is more useful than interleaved verbal reflection.

\paragraph{HotPotQA-32B.}
HotPotQA-32B is the most constrained setting: the gap is only $-1.13$\%, and 73.3\% of errors are genuine. Accordingly, the gate provides no measurable improvement (83.02\% $\rightarrow$ 82.90\%), while \textsc{critic} further reduces accuracy to 82.10\%. This directly supports our decomposition: when most errors arise from a genuine inability to synthesize information across passages, additional tool turns offer little help. The bottleneck lies in underlying reasoning capability rather than protocol execution.

\section{Variant-wise Robustness Under Semantic Distractors}
\label{app:variant_robustness}

\begin{figure*}[ht]
  \includegraphics[width=1\linewidth]{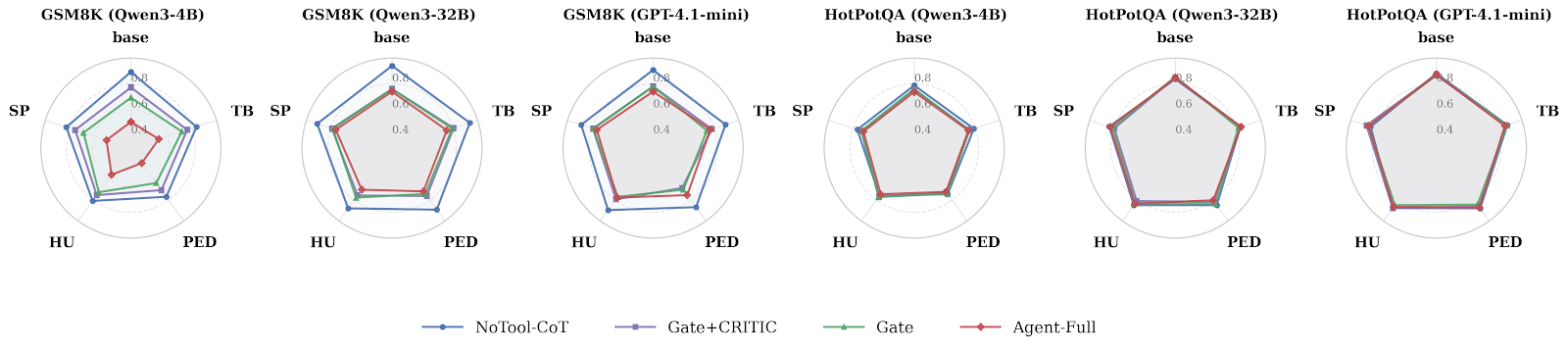}
\caption{Performance comparison of different reasoning strategies under semantic noise. The radar charts display the accuracy of Qwen3-4B, Qwen3-32B, and GPT-4.1-mini on GSM8K and HotPotQA. Each subplot contrasts four configurations (\textsc{CoT}, \textsc{Agent-Full}, \textsc{Gate}, \textsc{Gate+Critic}) evaluated across a clean \textit{base} set and four noisy variants (TB, PED, HU, SP).}
\label{fig:robust_radar}
\end{figure*}

Figure~\ref{fig:robust_radar} reports accuracy across the four semantic distractor categories (TB, SP, PED, HU) for Qwen3-4B and Qwen3-32B on GSM8K and HotPotQA. Across variants, \textsc{Gate} and \textsc{Gate+CRITIC} consistently improve over \textsc{Agent-Full}, indicating that gate-based control mitigates part of the degradation introduced by semantic distractors. The effect is most pronounced in fragile GSM8K settings; for example, on GSM8K-4B under PED, accuracy increases from 44.4\% to 70.4\% with \textsc{Gate+CRITIC}. This suggests that dynamic routing is not only an average-case improvement, but also a robustness mechanism against semantically relevant noise.

At the same time, \textsc{CoT} remains the strongest overall reference under distractors. Across variants, gated tool use narrows but does not eliminate the gap to \textsc{CoT}, suggesting that multi-turn tool interaction remains more vulnerable to distractor amplification than internal reasoning. This limitation is especially visible on HotPotQA-32B, where all methods remain tightly clustered, indicating that when the dominant bottleneck is genuine information synthesis rather than protocol execution, robustness gains from gate-based control are inherently limited.

Overall, the variant-wise results support the same qualitative conclusion as the main text: semantic distractors consistently expose the CoT--Tool gap, while gate-based control provides partial but not complete mitigation, with larger benefits in protocol-dominated settings than in capability-dominated ones.

\section{Prompt}
\label{Appendix:prompt}
\subsection{The Prompt for Generating Sem-Distractor}
In this section, we provide the prompts for generating distractors in Figure~\ref{fig:TB-Filler-Prompt}, Figure~\ref{fig:PED-Distractor-Prompt}, Figure~\ref{fig:HU-Hedging-Prompt} and Figure~\ref{fig:Paraphrase-Prompt}.

\begin{figure*}[t]
    \begin{AcademicBox}[\footnotesize The prompt used for generating Thematic Background (TB) filler]
    \textbf{Prompt: } \\
    "You are generating IN-DOMAIN background filler for a math word problem. These sentences should SOUND relevant to the same general topic/domain, but MUST NOT help solve the problem.\\
    \\
    Return STRICT JSON ONLY:\\
    \{\{\\
    \quad "topic": "<topic>",\\
    \quad "before": [<exactly \{BEFORE\_N\} short sentences>],\\
    \quad "after":  [<exactly \{AFTER\_N\} short sentences>]\\
    \}\}\\
    \\
    HARD RULES (must follow):\\
    1. Insert-only: write NEW sentences only; do NOT rewrite or paraphrase the evidence.\\
    2. Keep ONE coherent in-domain topic (same domain as the problem).\\
    3. NO numbers anywhere (no digits, no written numbers, no fractions).\\
    4. No math/solving strategy hints (forbidden: multiply, divide, sum, total, fraction, percent, per, each, equation, formula, calculate, compute).\\
    5. Do NOT use difference markers (forbidden: another, elsewhere, different, nearby, unrelated, separate, yesterday).\\
    6. Do NOT use hedging markers (forbidden: reportedly, claimed, might, possibly, about, around, likely, unverified).\\
    7. Do NOT mention the specific entities (names, places, dates) listed in: [\{CORE\_WORDS\}]. However, you MAY use general domain terms (e.g. 'clips', 'store', 'selling') to keep the topic relevant.\\
    8. Do NOT copy core numbers: [\{CORE\_NUMS\}]\\
    9. Each sentence must be short, factual, and self-contained."\\

    \vspace{-5pt} \hrule \vspace{4pt}
    \textbf{\textit{Output}} \\
    \textit{Input:} Problem core (evidence-centric view): \{Q\_CORE\_TEXT\}\\
    \textit{Output:}\\
    \end{AcademicBox}
    \vspace{-1em}
    \caption{The prompt used for generating Thematic Background (TB) Distractor.}
    \label{fig:TB-Filler-Prompt}
\end{figure*}

\begin{figure*}[t]
    \begin{AcademicBox}[\footnotesize The prompt used for generating Parrallel Entity Distractor (PED)]
    \textbf{Prompt: } \\
    "You are generating TOPIC-RELATED distractors that must be LOGICALLY EXCLUDED from the problem. They should share surface-level domain words, but clearly refer to a different person/time/place/event.\\
    \\
    Return STRICT JSON ONLY:\\
    \{\{\\
    \quad "topic": "<topic>",\\
    \quad "before": [<exactly \{BEFORE\_N\} short sentences>],\\
    \quad "after":  [<exactly \{AFTER\_N\} short sentences>]\\
    \}\}\\
    \\
    HARD RULES (must follow):\\
    1. Insert-only: write NEW sentences only; do NOT rewrite/paraphrase the evidence.\\
    2. Keep ONE coherent topic in the same domain.\\
    3. EACH sentence MUST include at least ONE difference marker: another / elsewhere / different / nearby / unrelated / separate / in a different / someone else / not this case.\\
    4. NO numbers anywhere.\\
    5. Must include 1–2 domain/topic hint words (same domain), but DO NOT restate evidence facts/relations.\\
    6. MUST NOT use hedging markers (forbidden: reportedly, claimed, might, possibly, about, around, likely).\\
    7. Use specific INVENTED names (e.g., 'Alice', 'Mr. Smith', 'The neighbor') instead of generic terms like 'someone' or 'another person' to make it sound natural.\\
    8. Do NOT copy core numbers: [\{CORE\_NUMS\}]\\
    9. Each sentence short and self-contained."\\

    \vspace{-5pt} \hrule \vspace{4pt}
    \textbf{\textit{Output}} \\
    \textit{Input:} Problem core (evidence-centric view): \{Q\_CORE\_TEXT\}\\
    \textit{Output:}\\
    \end{AcademicBox}
    \vspace{-1em}
    \caption{The prompt used for generating Parrallel Entity Distractor (PED).}
    \label{fig:PED-Distractor-Prompt}
\end{figure*}

\begin{figure*}[t]
    \begin{AcademicBox}[\footnotesize The prompt used for generating Hedged Uncertainty(HU) Distractor]
    \textbf{Prompt: } \\
    "You are generating TOPIC-RELATED statements that sound like evidence but are epistemically UNCERTAIN. These should not create contradictions that make the problem unsolvable.\\
    \\
    Return STRICT JSON ONLY:\\
    \{\{\\
    \quad "topic": "<topic>",\\
    \quad "before": [<exactly \{BEFORE\_N\} short sentences>],\\
    \quad "after":  [<exactly \{AFTER\_N\} short sentences>]\\
    \}\}\\
    \\
    HARD RULES (must follow):\\
    1. Insert-only: write NEW sentences only; do NOT rewrite/paraphrase the evidence.\\
    2. Keep ONE coherent topic in the same domain.\\
    3. MUST NOT include difference markers (forbidden: another, elsewhere, different, nearby, unrelated, separate, yesterday).\\
    4. NO numbers anywhere (avoid creating alternative exact facts).\\
    5. Do NOT assert the answer or options (forbidden: 'the answer is', 'correct answer', 'choose option', 'option X').\\
    6. You MAY mention the main entities (e.g. '\{CORE\_WORDS\}') to make the distractor sound like a valid premise, but the hedging marker must keep it uncertain.\\
    7. Do NOT copy core numbers: [\{CORE\_NUMS\}]\\
    8. Each sentence short and self-contained.\\
    9. EACH sentence MUST include at least ONE hedging marker: reportedly / claimed / it is said / some say / might / possibly / perhaps / around / about / likely / unverified / not confirmed.\\
    10. The tone should be speculative, vague, or rumor-based, distinct from the factual tone of the evidence."\\

    \vspace{-5pt} \hrule \vspace{4pt}
    \textbf{\textit{Output}} \\
    \textit{Input:} Problem core (evidence-centric view): \{Q\_CORE\_TEXT\}\\
    \textit{Output:}\\
    \end{AcademicBox}
    \vspace{-1em}
    \caption{The prompt used for generating Hedged Uncertainty(HU) Distractor.}
    \label{fig:HU-Hedging-Prompt}
\end{figure*}

\begin{figure*}[t]
    \begin{AcademicBox}[\footnotesize The prompt used for generating Semantic Paraphrase (SP) Distractor]
    \textbf{Prompt: } \\
    "You are generating SEMANTICS-PRESERVING PARAPHRASES of the EVIDENCE sentences in a word problem. Your output will be inserted BEFORE and AFTER the evidence as paraphrase fillers.\\
    \\
    Return STRICT JSON ONLY:\\
    \{\{\\
    \quad "topic": "<topic>",\\
    \quad "before": [<exactly \{BEFORE\_N\} short sentences>],\\
    \quad "after":  [<exactly \{AFTER\_N\} short sentences>]\\
    \}\}\\
    \\
    HARD RULES (must follow):\\
    1. One coherent topic: all sentences must be about the SAME scenario as the evidence.\\
    2. Paraphrase-only: every generated sentence MUST be a paraphrase of some [EVID] sentence content. Do NOT paraphrase [Q] (the question sentence). Do NOT add new facts, new events, or new entities.\\
    3. Syntactic Variation: You MUST change the sentence structure significantly (e.g., switch between active/passive voice, change word order). Do NOT simply replace one word; make the sentence LOOK different but MEAN the same.\\
    4. Numbers/values: you may mention numbers ONLY if they already appear in the evidence. Do NOT introduce any new numbers, ranges, approximations, or near-miss numbers. Keep exact values unchanged (no 'about', 'around', 'roughly', 'approximately').\\
    5. Keep relations unchanged: Preserve relational operators such as half / twice / remaining / difference / total. Do NOT change who did what, when, or the direction of a relation.\\
    6. No markers: Do NOT use difference markers (another / elsewhere / yesterday / different / nearby / unrelated / separate).\\
    7. No hedging markers: Do NOT use hedging markers (reportedly / claimed / might / possibly / about / around / likely / unverified).\\
    8. No solving hints: Do NOT include step-by-step solution language (multiply / divide / add / subtract / compute / calculate / equation / formula).\\
    9. Each sentence must be short, grammatical, and self-contained."\\

    \vspace{-5pt} \hrule \vspace{4pt}
    \textbf{\textit{Output}} \\
    \textit{Input:} Problem core (evidence-centric view): \{Q\_CORE\_TEXT\}\\
    \textit{Output:}\\
    \end{AcademicBox}
    \vspace{-1em}
    \caption{The prompt used for generating Semantic Paraphrase (SP) Distractor.}
    \label{fig:Paraphrase-Prompt}
\end{figure*}

\subsection{The Prompt for Intervention Framework}
In this section, we provide the prompts for Intervertion Framework in Figure~\ref{fig:CoT-Solver-Prompt} and Figure~\ref{fig:Tool-Solver-Prompt}.

\begin{figure*}[t]
    \begin{AcademicBox}[\footnotesize The prompt used for NoTool-CoT]
    \textbf{Prompt: } \\
    "You are a careful math solver. Read the numbered information chunks below. Some chunks are noise — use only the relevant ones as evidence.\\
    \\
    Rules:\\
    1. Think step-by-step in 'calc\_chain' FIRST before writing anything else.\\
    2. CRITICAL: Inside 'calc\_chain', enclose your final computed result in angle brackets and END with it, e.g. $<$42$>$.\\
    3. Then copy that exact value into 'final\_answer' (numeric only, no units).\\
    4. List chunk indices you actually used in 'evidence\_ids'.\\
    \\
    Return JSON with keys IN THIS ORDER: calc\_chain, evidence\_ids, final\_answer.\\
    \\
    CRITICAL MATH RULES:\\
    - 'half as many' $\rightarrow$ divide by 2; 'twice as much' $\rightarrow$ multiply by 2.\\
    - 'how much MORE does X need' = (what X needs) $-$ (what X already has + gifts).\\
    - 'X does Y to N people K times' $\rightarrow$ multiply by BOTH N and K.\\
    - Count EVERY multiplier in the sentence — don't miss any."\\

    \vspace{-5pt} \hrule \vspace{4pt}
    \textbf{\textit{Output}} \\
    \textit{Input:} \{NUMBERED\_CHUNKS\}\\
    \textit{Output:}\\
    \end{AcademicBox}
    \vspace{-1em}
    \caption{The prompt used for instructing the math solver to extract evidence, reason step-by-step, and format the final JSON output (NoTool-CoT).}
    \label{fig:CoT-Solver-Prompt}
\end{figure*}

\begin{figure*}[t]
    \begin{AcademicBox}[\footnotesize The prompt used for the tool-augmented solver]
    \textbf{Prompt: } \\
    "You are a math-solving agent with a 'calculate' function for arithmetic. You will see a question and numbered information chunks. Some chunks are noise — identify the real evidence.\\
    \\
    Strategy:\\
    1. Identify which chunks contain real evidence (specific numbers, clear facts) vs noise (hedged claims, unrelated info, different people/places/times).\\
    2. Extract relevant quantities from evidence chunks.\\
    3. Call the calculate function for ALL arithmetic — do NOT compute in your head.\\
    4. You may call calculate multiple times if needed (e.g. one step per operation).\\
    \\
    After all calculations, respond with JSON:\\
    \{\\
    \quad "evidence\_ids": [integer chunk indices that are real evidence],\\
    \quad "final\_answer": "the numeric answer (just the number, no units)",\\
    \quad "reasoning": "brief explanation of which chunks you used and why"\\
    \}\\
    \\
    CRITICAL RULES:\\
    - 'half as many' $\rightarrow$ divide by 2; 'twice as much' $\rightarrow$ multiply by 2.\\
    - 'how much MORE does X need' = (what X needs) $-$ (what X already has + gifts).\\
    - 'X does Y to N people K times' $\rightarrow$ multiply by BOTH N and K.\\
    - Count EVERY multiplier in the sentence — don't miss any."\\

    \vspace{-5pt} \hrule \vspace{4pt}
    \textbf{\textit{Output}} \\
    \textit{Input:} \{QUESTION\}, \{NUMBERED\_CHUNKS\}\\
    \textit{Output:}\\
    \end{AcademicBox}
    \vspace{-1em}
    \caption{The prompt used for instructing the tool-augmented math solver to identify evidence, utilize the calculate function, and format the final JSON output.}
    \label{fig:Tool-Solver-Prompt}
\end{figure*}

\subsection{The Prompt for Gate Mitigation}
In this section, we provide the prompts for Gate and Gate+Critic.

The Original Gate prompt is in Figure~\ref{fig:Gate-Verification-Prompt}.

The critic-style Gate prompt are in Figure~\ref{fig:Error-Correction-Prompt}, Figure~\ref{fig:Rederivation-Prompt}, Figure~\ref{fig:Repeated-Error-Prompt} and Figure~\ref{fig:Sense-Check-Prompt}.

The difference between original gate prompt and critic gate is the critic-style prompt tends to perform explicit verbal reasoning first, and then decides whether to re-trigger the tool.

\begin{figure*}[t]
    \begin{AcademicBox}[\footnotesize The prompt used for original gate]
    \textbf{Prompt: } \\
    "Your previous tool result is: \{prev\_output\}. Before finalizing, re-check the evidence chunks and verify:\\
    1) Did you use the correct numbers from evidence?\\
    2) Did you complete all required computation steps to the final answer?\\
    \\
    If anything is missing or incorrect, call calculate with a corrected and COMPLETE multi-step expression (do not repeat the same expression). If already complete and consistent, return final JSON only."\\

    \vspace{-5pt} \hrule \vspace{4pt}
    \textbf{\textit{Output}} \\
    \textit{Input:} Your previous tool result is: \{prev\_output\}\\
    \textit{Output:}\\
    \end{AcademicBox}
    \vspace{-1em}
    \caption{The prompt used for instructing the tool-augmented math solver to verify its evidence selection and calculation completeness before returning the final answer (Original Gate-style).}
    \label{fig:Gate-Verification-Prompt}
\end{figure*}

\begin{figure*}[t]
    \begin{AcademicBox}[\footnotesize The prompt used for error correction in thesolver]
    \textbf{Prompt: } \\
    "Calculator returned: \{prev\_output\}. Before calling calculate again, reason in words: which numbers from evidence do you need, and what is the correct step-by-step plan? Then call calculate with a CORRECTED expression (not the same one). Return final JSON after."\\

    \vspace{-5pt} \hrule \vspace{4pt}
    \textbf{\textit{Output}} \\
    \textit{Input:} Calculator returned: \{prev\_output\}\\
    \textit{Output:}\\
    \end{AcademicBox}
    \vspace{-1em}
    \caption{The prompt used for instructing the math solver to reflect on previous outputs, correct its reasoning, and formulate a new plan.}
    \label{fig:Error-Correction-Prompt}
\end{figure*}

\begin{figure*}[t]
    \begin{AcademicBox}[\footnotesize The prompt used for re-deriving the solution after a potentially incorrect tool result]
    \textbf{Prompt: } \\
    "Previous result (\{prev\_output\}) may be wrong. Re-derive the solution in words from evidence first, then call calculate with a revised expression. Do NOT reuse any previous expression. Return final JSON after."\\

    \vspace{-5pt} \hrule \vspace{4pt}
    \textbf{\textit{Output}} \\
    \textit{Input:} Previous result: \{prev\_output\}\\
    \textit{Output:}\\
    \end{AcademicBox}
    \vspace{-1em}
    \caption{The prompt used for instructing the tool-augmented math solver to re-derive its solution from the evidence and formulate a revised calculation when the previous result is suspected to be incorrect.}
    \label{fig:Rederivation-Prompt}
\end{figure*}

\begin{figure*}[t]
    \begin{AcademicBox}[\footnotesize The prompt used for handling repeated incorrect tool calls]
    \textbf{Prompt: } \\
    "You repeated: \{repeated\_expression\} (returned \{prev\_output\}). Explain in words why this is wrong and what should change, then call calculate with a DIFFERENT expression. Return final JSON after."\\

    \vspace{-5pt} \hrule \vspace{4pt}
    \textbf{\textit{Output}} \\
    \textit{Input:} You repeated: \{repeated\_expression\} (returned \{prev\_output\})\\
    \textit{Output:}\\
    \end{AcademicBox}
    \vspace{-1em}
    \caption{The prompt used for instructing the tool-augmented math solver to break out of a loop and correct a repeated incorrect calculation.}
    \label{fig:Repeated-Error-Prompt}
\end{figure*}

\begin{figure*}[t]
    \begin{AcademicBox}[\footnotesize The prompt used for step-by-step verification and sense-checking]
    \textbf{Prompt: } \\
    "Verify in words: re-read the question, list key numbers from evidence, check each arithmetic step. Does the answer make sense? If wrong, call calculate with a corrected expression. Return final JSON after."\\

    \vspace{-5pt} \hrule \vspace{4pt}
    \textbf{\textit{Output}} \\
    \textit{Input:} \{QUESTION\}, \{EVIDENCE\_CHUNKS\}, \{CURRENT\_ANSWER\}\\
    \textit{Output:}\\
    \end{AcademicBox}
    \vspace{-1em}
    \caption{The prompt used for instructing the math solver to verify its reasoning, check arithmetic steps against the evidence, and ensure the final answer makes logical sense.}
    \label{fig:Sense-Check-Prompt}
\end{figure*}

\section{Results in Noisy Condition}
\label{appendix:Results_in_Noisy_Condition}

In this section, we provide the full detailed results for GSM8K and HotPotQA across all conditions, variants, and metrics. The results are shown in Table~\ref{tab:full_results_gsm8k}, Table~\ref{tab:full_results_hotpotqa} and Table~\ref{tab:full_results_gpt41}.

\begin{table*}[htbp]
\renewcommand{\arraystretch}{1.1}
\setlength{\tabcolsep}{2pt}
\centering\small

\begin{tabular*}{\textwidth}{@{\extracolsep{\fill}} ll cccccc cccccc @{}}
\toprule
\multirow{2}{*}{\textbf{Condition}} & \multirow{2}{*}{\textbf{Metric}} & \multicolumn{6}{c}{\textbf{Qwen3-4B}} & \multicolumn{6}{c}{\textbf{Qwen3-32B}} \\
\cmidrule(lr){3-8} \cmidrule(lr){9-14}
& & Base & TB & PED & HU & SP & Overall & Base & TB & PED & HU & SP & Overall \\
\midrule
\multirow{3}{*}{NoTool-CoT} & Acc (\%) & 91.00 & 86.80 & 81.40 & 83.60 & 84.40 & 85.44 & 94.00 & 92.80 & 89.40 & 89.80 & 91.00 & 91.40 \\
& Ev-F1 (\%) & 89.78 & 89.72 & 89.99 & 78.15 & 19.44 & 73.41 & 92.48 & 93.35 & 95.24 & 85.89 & 35.00 & 80.39 \\
& AvgCalls & 0.00 & 0.00 & 0.00 & 0.00 & 0.00 & 0.00 & 0.00 & 0.00 & 0.00 & 0.00 & 0.00 & 0.00 \\
\midrule
\multirow{3}{*}{NoTool-FCStyle} & Acc (\%) & 85.80 & 85.80 & 82.20 & 86.60 & 83.80 & 84.84 & 80.00 & 80.40 & 75.40 & 76.40 & 80.60 & 78.56 \\
& Ev-F1 (\%) & 85.70 & 90.80 & 92.12 & 82.01 & 19.53 & 74.03 & 89.45 & 95.14 & 96.71 & 88.53 & 33.34 & 80.63 \\
& AvgCalls & 0.00 & 0.00 & 0.00 & 0.00 & 0.00 & 0.00 & 0.00 & 0.00 & 0.00 & 0.00 & 0.00 & 0.00 \\
\midrule
\multirow{3}{*}{Agent-NoopTool} & Acc (\%) & 34.40 & 29.40 & 29.80 & 30.40 & 29.20 & 30.64 & 53.40 & 50.40 & 50.40 & 49.20 & 51.20 & 50.92 \\
& Ev-F1 (\%) & 84.13 & 87.92 & 87.60 & 82.57 & 20.56 & 72.56 & 89.31 & 93.21 & 95.44 & 87.97 & 30.00 & 79.18 \\
& AvgCalls & 1.07 & 1.17 & 1.22 & 1.25 & 1.38 & 1.22 & 1.00 & 1.02 & 1.02 & 1.03 & 1.05 & 1.02 \\
\midrule
\multirow{3}{*}{Agent-Full} & Acc (\%) & 51.20 & 54.20 & 48.00 & 57.40 & 49.60 & 52.08 & 77.60 & 76.60 & 73.80 & 73.20 & 77.60 & 75.76 \\
& Ev-F1 (\%) & 84.78 & 87.81 & 88.77 & 83.11 & 19.45 & 72.78 & 89.01 & 93.02 & 95.26 & 87.42 & 30.03 & 78.95 \\
& AvgCalls & 1.26 & 1.35 & 1.39 & 1.43 & 1.59 & 1.41 & 1.03 & 1.04 & 1.04 & 1.04 & 1.12 & 1.05 \\
\cdashline{1-14}
\multirow{3}{*}{Agent-Max1Turn} & Acc (\%) & 47.40 & 50.80 & 44.40 & 53.00 & 43.00 & 47.72 & 78.60 & 76.20 & 73.20 & 73.80 & 77.60 & 75.88 \\
& Ev-F1 (\%) & 69.85 & 72.82 & 76.74 & 69.36 & 15.73 & 60.90 & 87.96 & 91.57 & 94.52 & 87.02 & 29.74 & 78.16 \\
& AvgCalls & 1.26 & 1.36 & 1.36 & 1.42 & 1.59 & 1.40 & 1.01 & 1.04 & 1.03 & 1.03 & 1.06 & 1.03 \\
\midrule
\multirow{3}{*}{Agent-OracleCalc} & Acc (\%) & 87.20 & 92.60 & 91.40 & 88.80 & 86.00 & 89.20 & 94.40 & 91.80 & 95.00 & 93.00 & 92.80 & 93.40 \\
& Ev-F1 (\%) & 84.09 & 88.08 & 89.24 & 83.47 & 20.13 & 73.00 & 89.17 & 93.38 & 95.18 & 87.75 & 30.19 & 79.13 \\
& AvgCalls & 1.20 & 1.28 & 1.34 & 1.37 & 1.55 & 1.35 & 1.00 & 1.02 & 1.02 & 1.03 & 1.10 & 1.03 \\
\midrule
\multirow{3}{*}{Agent-OracleEvid} & Acc (\%) & 50.00 & 53.00 & 53.00 & 53.20 & 53.20 & 52.48 & 78.80 & 79.80 & 80.00 & 80.00 & 80.00 & 79.72 \\
& Ev-F1 (\%) & 84.17 & 85.20 & 85.23 & 85.19 & 85.19 & 84.99 & 89.08 & 90.47 & 90.45 & 90.45 & 90.45 & 90.18 \\
& AvgCalls & 1.28 & 1.24 & 1.24 & 1.24 & 1.24 & 1.25 & 1.02 & 1.03 & 1.03 & 1.03 & 1.03 & 1.03 \\
\bottomrule
\end{tabular*}
\caption{Full detailed results for \textbf{GSM8K} across all conditions, variants, and metrics. Accuracy (Acc) and Evidence-F1 (Ev-F1) are in percentages (\%).}
\label{tab:full_results_gsm8k}

\end{table*}

\begin{table*}[htbp]
\renewcommand{\arraystretch}{1.1}
\setlength{\tabcolsep}{2pt}
\centering\small

\begin{tabular*}{\textwidth}{@{\extracolsep{\fill}} ll cccccc cccccc @{}}
\toprule
\multirow{2}{*}{\textbf{Condition}} & \multirow{2}{*}{\textbf{Metric}} & \multicolumn{6}{c}{\textbf{Qwen3-4B}} & \multicolumn{6}{c}{\textbf{Qwen3-32B}} \\
\cmidrule(lr){3-8} \cmidrule(lr){9-14}
& & Base & TB & PED & HU & SP & Overall & Base & TB & PED & HU & SP & Overall \\
\midrule
\multirow{3}{*}{NoTool-CoT} & Acc (\%) & 75.63 & 76.19 & 73.95 & 74.23 & 73.95 & 74.79 & 83.75 & 83.19 & 84.87 & 84.87 & 84.03 & 84.15 \\
& Ev-F1 (\%) & 59.72 & 60.03 & 60.70 & 56.07 & 28.62 & 53.03 & 56.86 & 58.51 & 59.10 & 55.41 & 25.05 & 50.99 \\
& AvgCalls & 0.00 & 0.00 & 0.00 & 0.00 & 0.00 & 0.00 & 0.00 & 0.00 & 0.00 & 0.00 & 0.00 & 0.00 \\
\midrule
\multirow{3}{*}{NoTool-FCStyle} & Acc (\%) & 71.15 & 71.99 & 69.75 & 69.75 & 71.99 & 70.92 & 85.15 & 84.87 & 82.91 & 83.19 & 83.75 & 83.98 \\
& Ev-F1 (\%) & 48.56 & 59.30 & 57.75 & 49.65 & 26.15 & 48.28 & 54.53 & 57.64 & 58.52 & 53.81 & 25.22 & 49.95 \\
& AvgCalls & 0.00 & 0.00 & 0.00 & 0.00 & 0.00 & 0.00 & 0.00 & 0.00 & 0.00 & 0.00 & 0.00 & 0.00 \\
\midrule
\multirow{3}{*}{Agent-NoopTool} & Acc (\%) & 49.30 & 59.10 & 52.66 & 57.42 & 64.99 & 56.69 & 84.03 & 81.79 & 80.39 & 80.95 & 83.19 & 82.07 \\
& Ev-F1 (\%) & 10.89 & 20.68 & 17.30 & 22.23 & 16.01 & 17.42 & 51.10 & 54.58 & 55.54 & 49.88 & 21.79 & 46.58 \\
& AvgCalls & 1.06 & 1.08 & 1.09 & 1.06 & 1.07 & 1.07 & 1.07 & 1.12 & 1.20 & 1.13 & 1.16 & 1.13 \\
\midrule
\multirow{3}{*}{Agent-Full} & Acc (\%) & 74.51 & 73.67 & 70.87 & 71.43 & 71.15 & 72.32 & 84.59 & 83.75 & 80.11 & 83.47 & 83.19 & 83.03 \\
& Ev-F1 (\%) & 47.01 & 52.19 & 49.44 & 46.25 & 32.74 & 45.53 & 48.34 & 50.70 & 51.19 & 48.89 & 27.86 & 45.40 \\
& AvgCalls & 1.06 & 1.08 & 1.09 & 1.06 & 1.08 & 1.07 & 1.06 & 1.11 & 1.17 & 1.14 & 1.15 & 1.13 \\
\cdashline{1-14}
\multirow{3}{*}{Agent-Max1Turn} & Acc (\%) & 71.99 & 70.03 & 68.91 & 69.47 & 69.47 & 69.97 & 84.03 & 84.03 & 80.67 & 83.47 & 83.75 & 83.19 \\
& Ev-F1 (\%) & 45.51 & 49.40 & 46.26 & 43.23 & 31.28 & 43.14 & 49.48 & 51.15 & 51.87 & 47.97 & 27.91 & 45.68 \\
& AvgCalls & 1.00 & 1.00 & 1.00 & 1.00 & 1.00 & 1.00 & 0.97 & 0.97 & 0.96 & 0.97 & 0.97 & 0.97 \\
\midrule
\multirow{3}{*}{Agent-OracleCalc} & Acc (\%) & 93.00 & 93.56 & 92.16 & 93.28 & 93.28 & 93.05 & 92.16 & 91.32 & 89.08 & 91.60 & 89.36 & 90.70 \\
& Ev-F1 (\%) & 38.60 & 44.04 & 45.09 & 39.62 & 21.78 & 37.82 & 50.71 & 53.83 & 53.94 & 50.32 & 22.18 & 46.19 \\
& AvgCalls & 1.07 & 1.08 & 1.09 & 1.06 & 1.07 & 1.07 & 1.04 & 1.10 & 1.19 & 1.12 & 1.14 & 1.12 \\
\midrule
\multirow{3}{*}{Agent-OracleEvid} & Acc (\%) & 74.51 & 73.95 & 74.51 & 74.51 & 74.51 & 74.40 & 84.87 & 83.75 & 85.15 & 85.15 & 85.15 & 84.82 \\
& Ev-F1 (\%) & 47.01 & 47.39 & 47.82 & 47.82 & 47.82 & 47.57 & 49.17 & 49.60 & 49.77 & 49.81 & 49.81 & 49.63 \\
& AvgCalls & 1.06 & 1.09 & 1.08 & 1.08 & 1.08 & 1.08 & 1.07 & 1.06 & 1.10 & 1.10 & 1.10 & 1.08 \\
\bottomrule
\end{tabular*}
\caption{Full detailed results for \textbf{HotPotQA} across all conditions, variants, and metrics. Accuracy (Acc) and Evidence-F1 (Ev-F1) are in percentages (\%).}
\label{tab:full_results_hotpotqa}
\end{table*}

\begin{table*}[htbp]
\renewcommand{\arraystretch}{1.1}
\setlength{\tabcolsep}{2pt}
\centering\small

\begin{tabular*}{\textwidth}{@{\extracolsep{\fill}} ll cccccc cccccc @{}}
\toprule
\multirow{2}{*}{\textbf{Condition}} & \multirow{2}{*}{\textbf{Metric}} & \multicolumn{6}{c}{\textbf{GSM8K}} & \multicolumn{6}{c}{\textbf{HotPotQA}} \\
\cmidrule(lr){3-8} \cmidrule(lr){9-14}
& & Base & TB & PED & HU & SP & Overall & Base & TB & PED & HU & SP & Overall \\
\midrule
\multirow{3}{*}{NoTool-CoT}
& Acc (\%)     & 93.20 & 91.00 & 89.20 & 90.40 & 89.80 & 90.72 & 87.96 & 87.96 & 87.96 & 87.68 & 83.75 & 87.06 \\
& Ev-F1 (\%)   & 90.99 & 94.44 & 95.62 & 86.86 & 25.91 & 78.76 & 49.47 & 52.26 & 53.10 & 50.07 & 22.48 & 45.48 \\
& AvgCalls     & 0.00  & 0.00  & 0.00  & 0.00  & 0.00  & 0.00  & 0.00  & 0.00  & 0.00  & 0.00  & 0.00  & 0.00 \\
\midrule
\multirow{3}{*}{NoTool-FCStyle}
& Acc (\%)     & 88.20 & 85.20 & 86.20 & 87.60 & 86.60 & 86.76 & 86.55 & 84.59 & 86.27 & 85.15 & 85.71 & 85.66 \\
& Ev-F1 (\%)   & 88.60 & 96.40 & 96.44 & 88.52 & 25.49 & 79.09 & 60.10 & 66.01 & 65.97 & 62.04 & 27.38 & 56.30 \\
& AvgCalls     & 0.00  & 0.00  & 0.00  & 0.00  & 0.00  & 0.00  & 0.00  & 0.00  & 0.00  & 0.00  & 0.00  & 0.00 \\
\midrule
\multirow{3}{*}{Agent-NoopTool}
& Acc (\%)     & 48.20 & 48.00 & 46.00 & 49.80 & 52.20 & 48.84 & 85.43 & 84.87 & 84.59 & 84.87 & 84.59 & 84.87 \\
& Ev-F1 (\%)   & 89.42 & 95.78 & 96.39 & 88.28 & 24.80 & 78.93 & 55.77 & 65.32 & 65.05 & 58.82 & 23.74 & 53.74 \\
& AvgCalls     & 1.18  & 1.11  & 1.11  & 1.05  & 1.65  & 1.22  & 1.17  & 1.55  & 1.68  & 1.62  & 1.45  & 1.49 \\
\midrule
\multirow{3}{*}{Agent-Full}
& Acc (\%)     & 75.40 & 76.00 & 76.60 & 77.80 & 77.20 & 76.60 & 87.11 & 85.99 & 87.11 & 86.55 & 85.43 & 86.44 \\
& Ev-F1 (\%)   & 89.77 & 95.52 & 95.27 & 88.83 & 23.73 & 78.63 & 58.21 & 60.25 & 60.37 & 56.75 & 33.76 & 53.87 \\
& AvgCalls     & 1.30  & 1.29  & 1.32  & 1.21  & 2.15  & 1.45  & 1.17  & 1.45  & 1.59  & 1.67  & 1.69  & 1.51 \\
\cdashline{1-14}
\multirow{3}{*}{Agent-Max1Turn}
& Acc (\%)     & 72.00 & 74.20 & 70.80 & 73.60 & 73.80 & 72.88 & 86.55 & 85.99 & 85.71 & 85.99 & 86.83 & 86.22 \\
& Ev-F1 (\%)   & 77.15 & 83.93 & 83.99 & 80.98 & 13.69 & 67.95 & 58.35 & 62.06 & 61.28 & 56.84 & 32.42 & 54.19 \\
& AvgCalls     & 1.22  & 1.23  & 1.20  & 1.13  & 2.10  & 1.37  & 1.00  & 1.00  & 1.00  & 1.00  & 1.00  & 1.00 \\
\midrule
\multirow{3}{*}{Agent-OracleCalc}
& Acc (\%)     & 84.20 & 85.40 & 82.60 & 82.60 & 76.40 & 82.24 & 91.88 & 92.16 & 91.04 & 92.72 & 90.76 & 91.71 \\
& Ev-F1 (\%)   & 89.47 & 95.62 & 95.68 & 88.16 & 25.17 & 78.82 & 54.14 & 62.84 & 62.91 & 58.29 & 23.04 & 52.24 \\
& AvgCalls     & 1.27  & 1.26  & 1.24  & 1.19  & 2.15  & 1.42  & 1.17  & 1.41  & 1.49  & 1.47  & 1.30  & 1.37 \\
\midrule
\multirow{3}{*}{Agent-OracleEvid}
& Acc (\%)     & 74.00 & 77.20 & 76.80 & 76.40 & 75.60 & 76.00 & 87.39 & 86.27 & 86.55 & 86.55 & 86.55 & 86.67 \\
& Ev-F1 (\%)   & 100.00 & 100.00 & 100.00 & 100.00 & 100.00 & 100.00 & 100.00 & 100.00 & 100.00 & 100.00 & 100.00 & 100.00 \\
& AvgCalls     & 1.28  & 1.28  & 1.28  & 1.28  & 1.26  & 1.28  & 1.22  & 1.23  & 1.20  & 1.17  & 1.23  & 1.21 \\
\bottomrule
\end{tabular*}
\caption{Full detailed results for \textbf{GPT-4.1-mini} across \textbf{GSM8K} and \textbf{HotPotQA} under all conditions, variants, and metrics. Accuracy (Acc) and Evidence-F1 (Ev-F1) are in percentages (\%). GSM8K uses exact-match accuracy. HotPotQA uses the same contains-match criterion as in Tables~2--7.}
\label{tab:full_results_gpt41}
\end{table*}

\section*{Limitations}

This document does not cover the content requirements for ACL or any
other specific venue.  Check the author instructions for
information on
maximum page lengths, the required ``Limitations'' section,
and so on.

\end{document}